\documentclass[review]{elsarticle}

\usepackage{lineno,hyperref}
\usepackage{amsmath}
\usepackage{enumitem}
\usepackage{diagbox}
\usepackage{booktabs}       
\usepackage{amsfonts}       
\usepackage{nicefrac}       
\usepackage{microtype}      
\usepackage{lipsum}
\usepackage{fancyhdr}       
\usepackage{graphicx}       
\graphicspath{{media/}}     
\usepackage{enumitem}
\usepackage{multirow}
\usepackage{graphicx}
\usepackage{caption}
\usepackage{amsmath}
\usepackage{pdfpages}
\usepackage{subcaption}
\usepackage{colortbl}
\usepackage{diagbox}
\usepackage{mathtools}
\usepackage{tikz}
\usepackage{algorithm}
\usepackage{bold-extra}
\usepackage{algpseudocode}
\algnewcommand{\LineComment}[1]{\State \(\triangleright\) #1}
\definecolor{blue-violet}{rgb}{0.54, 0.17, 0.89}
\definecolor{bluegray}{rgb}{0.4, 0.6, 0.8}
\definecolor{bleudefrance}{rgb}{0.19, 0.55, 0.91}
\definecolor{darkblue}{rgb}{0.0, 0.0, 0.55}
\definecolor{denim}{rgb}{0.08, 0.38, 0.74}
\definecolor{mediumblue}{rgb}{0.0, 0.0, 0.8}
\definecolor{naplesyellow}{rgb}{0.98, 0.85, 0.37}
\definecolor{carminered}{rgb}{1.0, 0.0, 0.22}
\definecolor{columbiablue}{rgb}{0.61, 0.87, 1.0}
\definecolor{ao(english)}{rgb}{0.0, 0.5, 0.0}
\definecolor{crimsonglory}{rgb}{0.75, 0.0, 0.2}
\definecolor{lightpastelpurple}{rgb}{0.69, 0.61, 0.85}
\definecolor{navajowhite}{rgb}{1.0, 0.87, 0.68}
\definecolor{tearose(rose)}{rgb}{0.96, 0.76, 0.76}
\definecolor{ruddypink}{rgb}{0.88, 0.56, 0.59}
\definecolor{lightgreen}{rgb}{0.56, 0.93, 0.56}
\definecolor{lightapricot}{rgb}{0.99, 0.84, 0.69}
\definecolor{pastelred}{rgb}{1.0, 0.41, 0.38}
\definecolor{periwinkle}{rgb}{0.8, 0.8, 1.0}
\definecolor{mediumchampagne}{rgb}{0.95, 0.9, 0.67}
\definecolor{dartmouthgreen}{rgb}{0.05, 0.5, 0.06}
\definecolor{xgreen}{rgb}{0.88, 0.95, 0.83}
\definecolor{airforceblue}{rgb}{0.36, 0.54, 0.66}
\definecolor{uared}{rgb}{0.85, 0.0, 0.3}
\definecolor{lightskyblue}{rgb}{0.53, 0.81, 0.98}
\definecolor{debianred}{rgb}{0.84, 0.04, 0.33}
\definecolor{skyblue}{rgb}{0.53, 0.81, 0.92}
\definecolor{red(ncs)}{rgb}{0.77, 0.01, 0.2}
\definecolor{blush}{rgb}{0.87, 0.36, 0.51}
\definecolor{non-photoblue}{rgb}{0.64, 0.87, 0.93}
\definecolor{lightcornflowerblue}{rgb}{0.6, 0.81, 0.93}
\definecolor{dodgerblue}{rgb}{0.12, 0.56, 1.0}
\definecolor{cadmiumgreen}{rgb}{0.0, 0.42, 0.24}
\definecolor{dollarbill}{rgb}{0.52, 0.73, 0.4}
\definecolor{palegoldenrod}{rgb}{0.93, 0.91, 0.67}
\definecolor{deeppink}{rgb}{1.0, 0.08, 0.58}
\definecolor{lightmauve}{rgb}{0.86, 0.82, 1.0}
\definecolor{darkgoldenrod}{rgb}{0.72, 0.53, 0.04}
\definecolor{deepcerise}{rgb}{0.85, 0.2, 0.53}
\definecolor{darkgreen}{rgb}{0.0, 0.2, 0.13}
\newcommand\DenseLinkSearch{\textsc{DenseLinkSearch}}
\newcommand\OpenI{\textcolor{airforceblue}{{\textsc{Open}}}\circled{\textbf{\large \textit{i}}}}
\newcommand\resnet{{ResNet50}}
\newcommand\vitbase{{ViT-Base}}
\newcommand\vitlarge{{ViT-Large}}
\newcommand\covertree{{Cover Tree}}

\newcommand\vptree{{VP Tree}}
\newcommand\kdtree{{KD Tree}}
\newcommand\pcatree{{PCA Tree}}
\newcommand\rtree{{R Tree}}
\newcommand\balltree{{Ball Tree}}
\newcommand\mtree{{M Tree}}
\newcommand\vithuge{{ViT-Huge}}
\newcommand*\circled[1]{\tikz[baseline=(char.base)]{
            \node[shape=circle,draw,fill={rgb:red,4;green,2;yellow,1}, scale=0.75,inner sep=2pt] (char) {#1};}}

\errorcontextlines\maxdimen

\makeatletter
\newcommand*{\algrule}[1][\algorithmicindent]{\makebox[#1][l]{\hspace*{.5em}\thealgruleextra\vrule height \thealgruleheight depth \thealgruledepth}}%
\newcommand*{\thealgruleextra}{}
\newcommand*{\thealgruleheight}{.75\baselineskip}
\newcommand*{\thealgruledepth}{.25\baselineskip}

\newcount\ALG@printindent@tempcnta
\def\ALG@printindent{%
    \ifnum \theALG@nested>0
        \ifx\ALG@text\ALG@x@notext
        \else
            \unskip
            \addvspace{-1pt}
            \ALG@printindent@tempcnta=1
            \loop
                \algrule[\csname ALG@ind@\the\ALG@printindent@tempcnta\endcsname]%
                \advance \ALG@printindent@tempcnta 1
            \ifnum \ALG@printindent@tempcnta<\numexpr\theALG@nested+1\relax
            \repeat
        \fi
    \fi
    }%
\usepackage{etoolbox}
\patchcmd{\ALG@doentity}{\noindent\hskip\ALG@tlm}{\ALG@printindent}{}{\errmessage{failed to patch}}
\makeatother

\newbox\statebox
\newcommand{\myState}[1]{%
    \setbox\statebox=\vbox{#1}%
    \edef\thealgruleheight{\dimexpr \the\ht\statebox+1pt\relax}%
    \edef\thealgruledepth{\dimexpr \the\dp\statebox+1pt\relax}%
    \ifdim\thealgruleheight<.75\baselineskip
        \def\thealgruleheight{\dimexpr .75\baselineskip+1pt\relax}%
    \fi
    \ifdim\thealgruledepth<.25\baselineskip
        \def\thealgruledepth{\dimexpr .25\baselineskip+1pt\relax}%
    \fi
    \State #1%
    \def\thealgruleheight{\dimexpr .75\baselineskip+1pt\relax}%
    \def\thealgruledepth{\dimexpr .25\baselineskip+1pt\relax}%
}

\DeclareMathOperator*{\argmin}{arg\,min}

\modulolinenumbers[5]

\journal{Artificial Intelligence In Medicine}

\begin{document}
\nolinenumbers
\begin{frontmatter}

\title{Medical Image Retrieval via Nearest Neighbor Search on Pre-trained Image Features }


 \author{Deepak Gupta\corref{cor1}}
  \ead{deepak.gupta@nih.gov}
  \cortext[cor1]{Corresponding author}
    \author{Russell Loane}
    \ead{russellloane@gmail.com}
    \author{Soumya Gayen\corref{cor2}}
 \ead{soumya.gayen@nih.gov}
     \author{Dina Demner-Fushman\corref{cor2}}
 \ead{ddemner@mail.nih.gov}
    \address{Lister Hill National Center for Biomedical Communications \\
National Library of Medicine, National Institutes of Health \\
Bethesda, MD, USA
}




\begin{abstract}
Nearest neighbor search (NNS) aims to locate the points in high-dimensional space that is closest to the query point. The brute-force approach for finding the nearest neighbor becomes computationally infeasible when the number of points is large. The NNS has multiple applications in medicine, such as searching large medical imaging databases, disease classification, and diagnosis. With a focus on medical imaging, this paper proposes \DenseLinkSearch{} an effective and efficient algorithm that searches and retrieves the relevant images from heterogeneous sources of medical images. Towards this, given a  medical database, the proposed algorithm builds an index that consists of pre-computed links of each point in the database. The search algorithm utilizes the index to efficiently traverse the database in search of the nearest neighbor. We extensively tested the proposed NNS approach and compared the performance with state-of-the-art NNS approaches on benchmark datasets and our created medical image datasets. The proposed approach outperformed the existing approaches in terms of retrieving accurate neighbors and retrieval speed. We also explore the role of medical image feature representation in content-based medical image retrieval tasks. We propose a Transformer-based feature representation technique that outperformed the existing pre-trained Transformer-based approaches on CLEF 2011 medical image retrieval task.
The source code and datasets of our experiments are available at \url{https://github.com/deepaknlp/DLS}.
\end{abstract}

\end{frontmatter}

\nolinenumbers

\section{Introduction}
Over the past few decades, medical imaging has significantly improved healthcare services. Medical imaging helps to save lives, increase life expectancy, lower mortality rates, reduce the need for exploratory surgery, and shorten hospital stays. With medical imaging, the physician makes better medical decisions regarding diagnosis and treatment. Medical imaging procedures are non-invasive and painless and often do not necessitate any particular preparation beforehand. With the growing demand for medical imaging, the workload of radiologists has increased significantly over the past decades. 
Mayo Clinic has observed a ten-fold increase in the demand for radiology imaging from just over $9$ million in $1999$ to more than $94$ million in $2010$ \citep{mcdonald2015effects}. To meet the growing demand, radiologists must process one image every three to four seconds \citep{mcdonald2015effects}. Consequently, the increase in workload may lead to the incorrect interpretation of the radiology images and compromise the quality and safety of patient care. 

The recent advancement in the Artificial intelligence (AI) fields of computer vision and machine learning has the potential to quickly interpret and analyze different forms of medical images \citep{lambin2012radiomics,gupta2021hierarchical,yu2020cross} and videos \citep{gupta2022dataset,gupta-demner-fushman-2022-overview}. Content-based image retrieval (CBIR) is one of the key tasks in analyzing medical images. It involves indexing the large-scale medical-image datasets and retrieving visually similar images from the existing datasets. With an efficient CBIR system, one can browse, search, and retrieve from the databases images that are visually similar to the query image. 

CBIR systems are used to support cancer diagnosis  \citep{wei2009microcalcification,bressan2019breast}, diagnosis of infectious diseases \citep{zhong2021deep} and analyze the central nervous system \citep{mesbah2015hashing,conjeti2016neuron,li2018large}, biomedical image archive \citep{antani2004content}, malaria parasite detection \citep{khan2011content,rajaraman2018pre,kassim2020clustering}. 
Given the growing size of the medical imaging databases, efficiently finding the relevant images is still an important issue to address. Consider a large-scale medical imaging database with hundreds of thousands to millions of medical images, in which each image is represented by high-dimensional (thousands of features) dense vectors. Searching over the millions of images in such high-dimensional space requires an efficient search. The features used to represent the image are another key aspect that affects the image search results. Image features with reduced expressive ability often fail to discriminate the images with the near-similar visual appearance. The role of image features becomes more prominent with image search applications that search over millions of images and demand a higher degree of precision. To address the aforementioned challenges, we focus on developing an algorithm that can efficiently search over millions of medical images. We also examine the role of image features in obtaining relevant and similar images from large-scale medical imaging datasets.

This study presents \DenseLinkSearch{} an efficient algorithm to search and retrieve the relevant images from the heterogeneous sources of medical images and nearest neighbor search benchmark datasets. We first index the feature vectors of the images. The indexing produces a graph with feature vectors as vertices and euclidean distance between the  endpoints vectors as edges. In the literature, the tree-based data structure has been used to build indexes to speed up search retrieval. \citet{beygelzimer2006cover} proposed \covertree{} that was specifically designed to facilitate the speed up of the nearest neighbor search by efficiently building the index. 
We compare our proposed \DenseLinkSearch{} with the existing tree-based and approximate nearest neighbor approaches and provide a detailed quantitative analysis.

To evaluate the proposed \DenseLinkSearch{} algorithm, we collected $12,851,263$ medical images from the \OpenI{}\footnote{\url{https://openi.nlm.nih.gov/}} biomedical search engine. 
We extend our experiments on $11$ benchmarked NNS datasets (Artificial, Faces, Corel, MNIST, FMNIST, TinyImages, CovType, Twitter, YearPred, SIFT, and GIST). The experimental results show that our proposed \DenseLinkSearch{} is more efficient and accurate in finding the nearest neighbors in comparison to the existing approaches.

We summarize the contributions of our study as follows:
\begin{enumerate}[noitemsep]
    \item We devise a robust nearest neighbor search algorithm \DenseLinkSearch{} to efficiently search large-scale datasets in which the data points are often represented by the high dimension vectors. To perform the search, we develop an indexing technique that processes the dataset and builds a graph to store the link information of each data point present in the dataset. The created graph in the form of an index is used to quickly scan over the millions of data points in search of the nearest neighbors of the query data point. 
    \item We also perform an extensive study on the role of features that are used to represent medical images in the dataset. To assess the effectiveness of the features in retrieving the relevant images, we explore multiple deep neural-based features such as  ResNet, ViT, and ConvNeXt  and analyze their effectiveness in accurately representing the images in high-dimensional spaces.
    \item We demonstrate the effectiveness of our proposed \DenseLinkSearch{} on newly created \OpenI{} medical imaging datasets and eleven other benchmarked NNS datasets. The results show that our proposed NNS technique accurately searches the nearest neighbor orders of magnitude faster than any comparable algorithm. 
    
\end{enumerate}
\section{Related Work} \label{section:related-work}
\subsection{Content-based Image Retrieval} 
Content-based image retrieval focuses on retrieving images by considering the visual content of the image, such as color, texture, shape, size, intensity, location, etc. For the instance of medical image retrieval,
\citet{xue2008web} introduced the CervigramFinder system that operates on cervicographic images and aims to find similar images in the database as per the user-defined region. The system extracted color, texture, and size as the visual features. \citet{antani2007interfacing} developed SPIRS-IRMA that combines the capability of IRMA \citep{lehmann2004content} system (global image data) and SPIRS \citep{hsu2007spirs} system (local region-of-interest image data) to facilitate retrieval based not only on the whole image but also on local image features so that users can retrieve images that are not only similar in terms of their overall appearance but also similar in terms of the pathology that is displayed locally. \citet{depeursinge20113d} proposed a 3D localization system based on lung anatomy that is used to localize low-level features used for CBIR. The image retrieval task of the Conference and Labs of the Evaluation Forum (ImageCLEF) has organized multiple medical image retrieval tasks \citep{clough2004clef,muller2009overview,kalpathy2011overview,muller2012overview} from the year 2004 to 2013. ImageCLEF has provided a venue for the researcher to present their findings and engage in head-to-head comparisons of the efficiency of their medical image retrieval strategies. Over the years, the participants at ImageCLEF made use of a diverse selection of local and global textural features. These included the Tamura features: coarseness, contrast, directionality, line-likeness, regularity, and roughness. Multiple filters such as Gabor, Haar, and Gaussian filters have been used to generate a diverse set of visual features. The visual features \citep{kalpathy2015evaluating} for medical image retrieval are also generated using Haralick's co-occurrence matrix and fractal dimensions. 

\citet{rahman2008medical} proposed a content-based image retrieval framework that deals with the diverse collections of medical images of different modalities, anatomical regions, acquisition views, and biological systems. They extracted the low-level image features such as MPEG (Moving Picture Experts Group)-7 based Edge Histogram Descriptor (EHD) and Color Layout Descriptor (CLD) to represent the images. Further, \citet{rahman2011learning} presents an image 
retrieval framework based on image filtering and image similarity fusion. The framework utilizes the support vector machine (SVM) \citep{cortes1995support}  to predict the category of query images and images stored in the database. In this way, the irrelevant images are filtered out, which leads to reduced search space for image similarity matching. A three-stage approach for human brain magnetic resonance image retrieval was introduced by \citet{nazari2010cbir}. In the first stage, the gray level co-occurrence matrix (GLCM) \citep{haralick1979statistical}
was constructed thereafter, the image features were extracted by computing Features Energy, Entropy, Contrast, Inverse Difference Moment, Variance, Sum Average, Sum Entropy, Sum Variance, Difference Variance, Difference Entropy, and Information measure of correlation. Principal component analysis (PCA) was used for feature reduction in the second stage. An SVM classifier was used in the last stage to perform decision-making. 

With the success of the convolution neural network (CNN) for image classification, CNN-based pre-trained models \citep{simonyan2014very,he2016deep,huang2017densely,szegedy2016rethinking,szegedy2015going} became the de-facto architecture for image classification, feature extraction, and analysis. \citet{qayyum2017medical} proposed a deep learning-based framework for medical image retrieval tasks. A deep convolutional neural network was trained for the medical image classification, and the trained model was used to extract the image features. \citet{cao2014medical} developed a deep Boltzmann machine-based multimodal learning model for fusion of the visual and textual information. The proposed multimodal approach enabled searches of the most relevant images for a given image query.

Off-the-shelf pre-trained language models were used to extract the image features for the open domain \citep{chen2021deep}. The straightforward approach \citep{sharif2014cnn,gong2014multi,babenko2014neural} is to extract the image features from the last fully-connected layer; however, it may include irrelevant patterns or background clutter. In another strategy \citep{yue2015exploiting,razavian2016visual,lou2018multi}, the image features are extracted from convolutional layers that preserve more structural details. In order to extract global and local features for the images, the layer-level \citep{do2017embedding,cao2020unifying,yu2017exploiting,zhang2019effective} feature fusion mechanism has been adapted to complement each feature for the image retrieval task.

\subsection{Nearest Neighbor Search}
In the course of the last several decades, numerous optimization strategies that are aimed at speeding up the nearest neighbor search have been presented. The tree-based methods have been widely used to speed up the NNS process. In the tree-based approach, a tree-like data structure is used to organize the data points in such a way that it can be efficiently traversed in search of the nearest neighbors for the input query. In general, once the tree structure has been built, the triangle inequality is used to filter the nodes of the tree that can not be the nearest neighbor, thereby reducing the computation and speeding-up the search process. \citet{friedman1977algorithm} proposed the first space-partitioning tree \kdtree{}  that uses a depth-first tree traversal technique, followed by backtracking to locate the nearest neighbor in logarithmic time. At each node of \kdtree{}, $d$-dimensional data points are recursively partitioned into two sets by splitting along one dimension of the data. The split value is often determined to be the median value along the dimension being split. This leads to the points being evenly distributed over the axis-aligned hyper-rectangles. Choosing the split value and split dimensions are the key challenges in \kdtree{}. To alleviate these issues, \balltree{} \citep{fukunaga1975branch,omohundro1989five} was proposed that considered the hyper-spheres instead of hyper-rectangles that form a cluster of the data points in high-dimensional spaces. The \balltree{} computes the centroid of the whole data set, which is then used to recursively partition the data set into two subgroups.
It uses triangle inequality to prune the ball and all data points within the ball while searching for nearest neighbors. Other tree-based approaches such as \pcatree{}, \citep{sproull1991refinements}, \vptree{} \citep{yianilos1993data}, \mtree{} \citep{ciaccia1997m}, \rtree{} \citep {kamel1993hilbert}, and \covertree{} \citep{beygelzimer2006cover} have been introduced in the literature of nearest neighbour search. Most of the tree-based methods for nearest-neighbor search are often well-suited for low dimensional data points, however, they perform poorly in high dimensional spaces. 

\citet{wang2011fast} observed that the performance of tree-based methods is considered to be satisfactory if the search for nearest neighbors requires only a few nodes at each level of the tree. However, in the case of high dimensional data, these approaches lose their effectiveness as the histograms of distances and 1-Lipschitz function values become concentrated \citep{pestov2012indexability,boytsov2013learning}. In this case,  indexes built with a clustering-based partition technique seem to perform better than the tree-based indexes. By following the clustering-based partition scheme, numerous approaches \citep{prerau2000unsupervised,wang2011fast,almalawi2015k} have been proposed to find the nearest neighbors in high dimensional spaces. In clustering-based indexes, the data points form multiple clusters. While searching for the nearest neighbors, the triangle inequality can be applied to prune the clusters that can't hold the nearest neighbors.

In another line of research, lower bound-based methods for efficient nearest neighbor search have been proposed. The key idea of the lower bound-based methods is to reduce the distance computation between the query and the candidate data points, which leads to an efficient NNS. \citet{liu2018exploiting} proposed two lower bounds: progressive lower bound (PLB) and statistical lower bound (SLB), that aim to reduce the distance computation and accelerate the approximate NNS of the HNSW indexing method \citep{malkov2014approximate}. \citet{hwang2012fast} consider the mean and variance of data points to derive the lower bounds that significantly reduce the distance computations. Further, \citet{hwang2018product} introduced product quantized translation that aims to eliminate nearest neighbor candidates effectively using their euclidean distance lower bounds in nonlinear embedded spaces. \citet{jeong2006effective,li2018efficient} also utilized the lower bounds strategy to reduce the distance computations. Recently, \citet{zhang2022accelerating} introduced the concept of block vectors based on lower bounds that reduce the expensive distance computation. Further, they designed a multilevel lower bound that computes the lower bound step-by-step and makes use of the multistep filtering technique to speed up the search further.

\section{Proposed Approach}
\subsection{Background}
In the nearest neighbor search, given a set of $\mathcal{N}$ data points $\mathcal{D}=\{v_1, v_2, \ldots, v_{\mathcal{N}}\}$ where $v_i \in \mathcal{R}^{d}$, and a distance metric $dist(v_i, v_j)$ for points $v_i$ and $v_j$, for any given query point $q \in \mathcal{R}^{d}$, the goal is to find the nearest point $v^{*}$ in the data points such that:
\begin{equation}
    \label{eq:nns}
    v^* = \argmin_{v_i \in \mathcal{D}} dist(v_i, q)
\end{equation}
A variant of NNS is the $k$-nearest neighbors (kNN) problem, which aims to find the $k$ nearest points in $\mathcal{D}$ to query $q$, where $k$ is a constant. A brute-force algorithm requires computing the distance between a query point $q$ and each data point in $\mathcal{D}$, resulting in $\mathcal{O}(\mathcal{N})$ time. In applications where the number of points is large, and each data point is high dimensional, it is not computationally feasible to use the brute-force algorithm. Therefore, our goal is to process the data points in advance, which can reduce the computational complexity and quickly find the nearest neighbors of the queries.

\subsection{Nearest Neighbor Search}
We propose an effective approach to finding the kNN in high-dimensional vector space. The approach deals with building the index and finding the nearest neighbors to the given query with a time-efficient \DenseLinkSearch{} algorithm. In this section, we describe the indexing algorithm and \DenseLinkSearch{} algorithm in detail.
\subsubsection{Summary of the Approach}
Given a high-dimensional dataset, our proposed approach first builds the index considering all the data points from the dataset. Formally, the indexing algorithm builds the \textit{descend} and \textit{spread} links (to be introduced shortly) for each data point by considering the $\mathcal{K}_{index}$ nearest neighbors to them. We use the notation $\mathcal{K}_{index}$  to indicate the number of nearest neighbors embedded in the index. In the indexing process, each vertex\footnote{We use the term `vertex,' `vector', and `data point' interchangeably in the paper.} forms links to the other vertices, making a cluster of the links that appear to be dense links. While searching for the  neighbors nearest to a query after the index is built, \DenseLinkSearch{} algorithm follows the two-stage approach. In the first stage, it follows the descend links of the closest indexed vector to the query vector, and in the second stage, it follows the spread links of the closest indexed vector to surround the query vector. 
While searching the neighbors, the algorithm spends the majority of the calculation time in the spread stage to find the $\mathcal{K}_{search}$ nearest neighbors. We use the notation $\mathcal{K}_{search}$ to indicate the number of nearest neighbors to the query image that has to be searched in the dataset. 
\begin{figure}
    \centering
    \includegraphics[width=\linewidth]{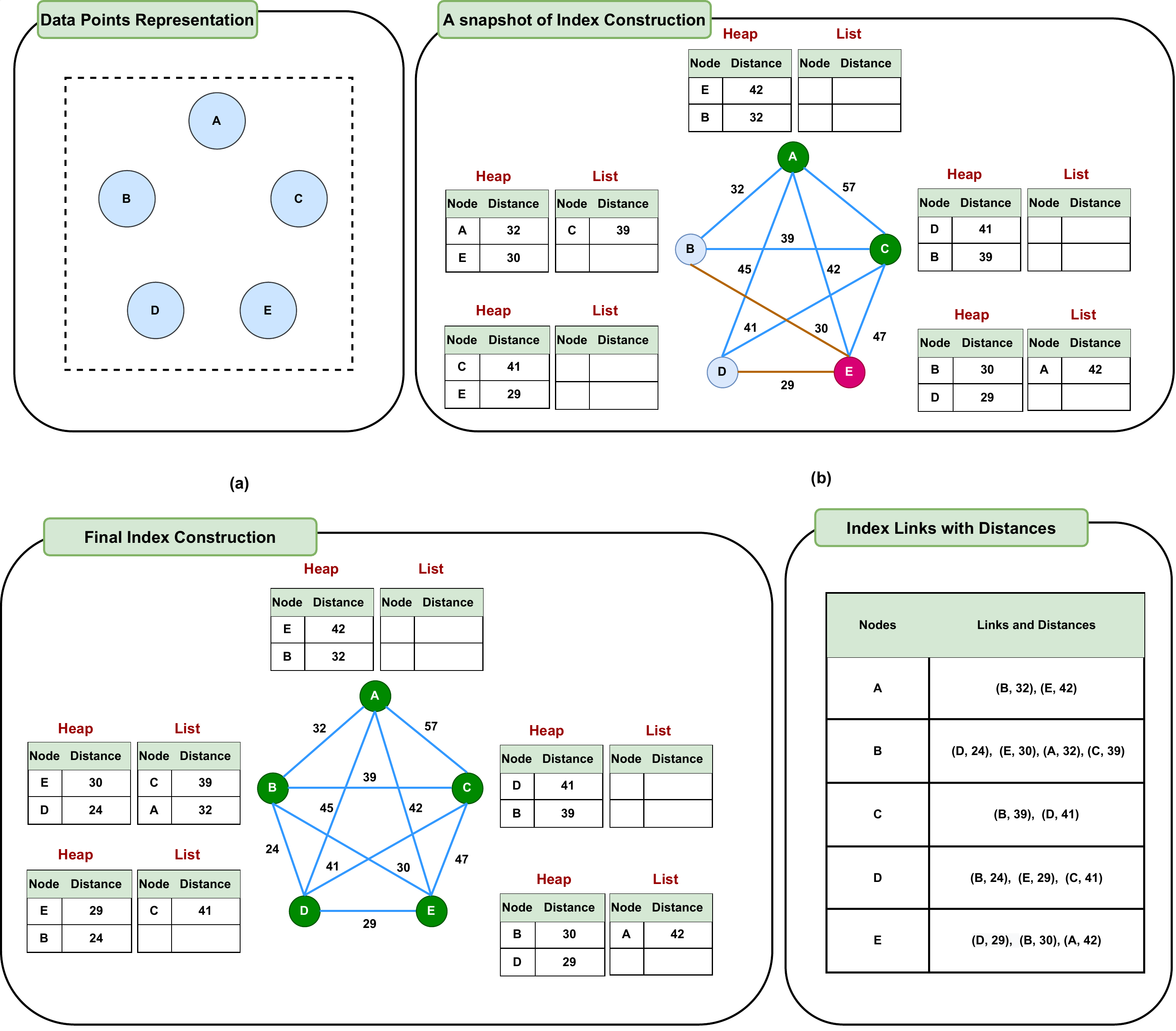}
    \caption{Illustration of the indexing algorithm. Subfigure \textbf{(a)} demonstrates the image vector representation in high-dimensional space. The indexing algorithm considers each vector representation as a vertex and the distance between them as the edge of the index graph. The indexing starts with a random vertex and calls the vertex a node. At any point of time during index construction, a vector representation that has not become a node of the index graph is shown as \protect\begin{tikz}[baseline=-1mm] \protect\draw[fill=columbiablue] circle [radius=0.15] node {$\textcolor{white}{\mathbf{}}$}; \protect\end{tikz}. The algorithm initializes a heap data structure of size $\mathcal{K}_{index}=2$ and a list. The heap holds links to the nearest vectors known so far. The list holds links to other vectors that consider this vector near. The top link in a heap defines the neighborhood radius, also known as the near distance. Subfigure \textbf{(b)} represents a snapshot of the index after the data point \textbf{E} became a node, and its heap and list are updated based on distances between the node and its neighbors. The existing nodes and edges are shown as \protect\begin{tikz}[baseline=-1mm] \protect\draw[fill=dartmouthgreen] circle [radius=0.15] node {$\textcolor{white}{\mathbf{}}$}; \protect\end{tikz}
   and \protect\begin{tikzpicture}[baseline=-1mm] \protect\draw[dodgerblue] (0,0) -- (0.5,0); \protect\end{tikzpicture} respectively. The recently processed node \textbf{E} and added edges are shown as \protect\begin{tikz}[baseline=-1mm] \protect\draw[fill=deepcerise] circle [radius=0.15] node {$\textcolor{white}{\mathbf{}}$}; \protect\end{tikz} and \protect\begin{tikzpicture}[baseline=-1mm] \protect\draw[darkgoldenrod] (0,0) -- (0.5,0); \protect\end{tikzpicture} respectively. Subfigure \textbf{(c)} demonstrates each node's heap and list items and the index graph with nodes and edges after the index is built. The created links and distances are shown in subfigure \textbf{(d)}. The final index contains the descend and spread links for each node in the dataset. The descend links contain the endpoints vertices and distances from the heap just before the vertex becomes a node while building the indexes. Similarly, the spread links of a given vertex store the endpoints vertices and distances from the heap and list once the algorithm ends.
    }
    \label{fig:indexing}
\end{figure}
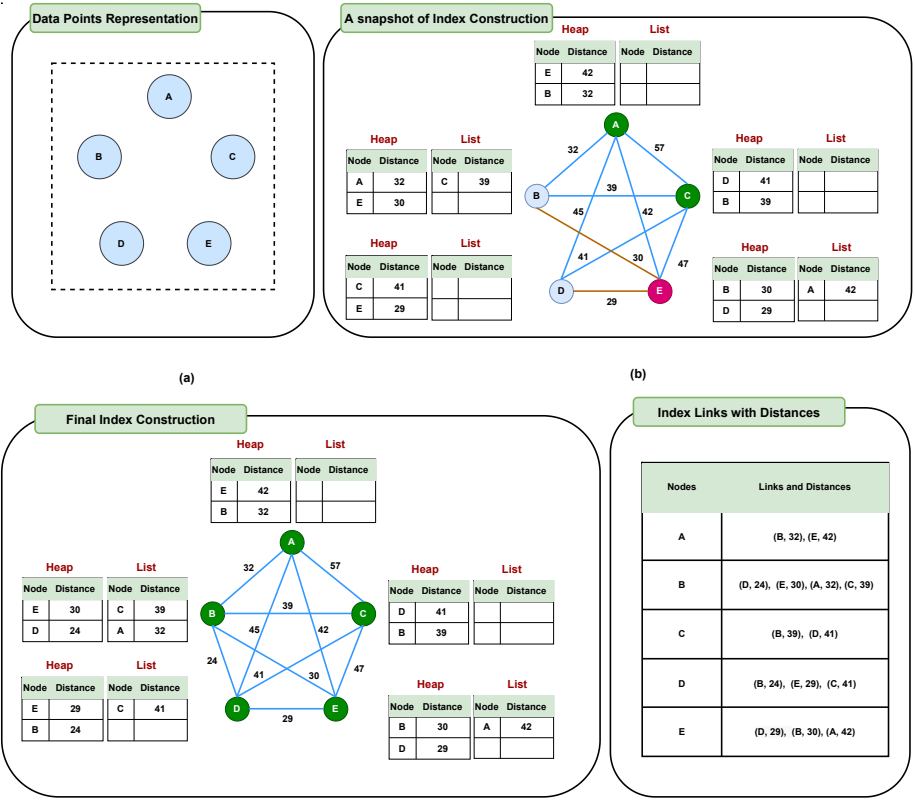

\subsubsection{Indexing Algorithm}
Given a dataset $\mathcal{D}$ having $\mathcal{N}$ vectors, the algorithm builds the index by considering the $\mathcal{K}_{index}$ nearest neighbors for each vector in the data. The indexing algorithm constructs an index graph $\mathcal{G}$ where the vertices of the graph are the vectors, and the edges between them hold the Euclidean distances between the two vertices of the graph. The vectors are added to the index one by one. Once the vector is added to the index, it is called a node (vertex), so all nodes are vectors. 
During indexing, each vector has a spherical neighborhood that encloses neighboring vectors. Neighborhoods contract from infinity (the entire set of data points) at the start of indexing to just the nearest neighbors at the end. When a vector becomes a node, its final neighborhood (the $\mathcal{K}_{index}$ nearest neighbors) is calculated, so its neighborhood shrinks to a minimum.  In this process, links are created to neighbors, often causing their neighborhoods to shrink a little as well. Initially, all neighborhoods are huge and overlap, so links are created to all nodes. Later on, neighborhoods overlap with fewer and fewer other neighborhoods, creating fewer links. Shrinking neighborhoods allows indexing to occur with much less than $\mathcal{O}(\mathcal{N}^2)$ distance calculations.

\begin{table}[]
\resizebox{\textwidth}{!}{%
\begin{tabular}{l|l}
\hline
\textbf{Variable}               & \textbf{Description} \\ \hline
$\mathcal{R}_v$    & Neighborhood radius, also known as near distance.         \\ \hline
$\mathcal{C}_v$     & Distance to closest indexed vector.        \\ \hline
$\mathcal{H}_v$     &\begin{tabular}[c]{@{}l@{}}Heap to hold the links to the near neighbors of vector $v$. The top of the heap $\mathcal{H}_v$ is the link \\ to the furthest of the near neighbor vector.\end{tabular}    \\ \hline
$\mathcal{L}_v$     &\begin{tabular}[c]{@{}l@{}}List to hold the links to the far neighbors of vector $v$. This list hold links to  other vectors\\ that consider this vector near.\end{tabular}    \\ \hline
$\mathcal{I}_v$     & List to hold finished index links for vector $v$.          \\ \hline
\end{tabular}%
}
\caption{Description of the variables for each vector $v \in \mathcal{D}$ .}
\label{tab:vector_variables}
\end{table}

The proposed indexing algorithm has the following steps to build the index:
\begin{enumerate}
    \item \textbf{Initialization}: 
    For each vector $v$ in the dataset $\mathcal{D}$, we initialize the variables listed in Table \ref{tab:vector_variables}. We also initialize a global max-heap $\mathcal{H}$ of size $\mathcal{N}$ that stores the distance of the vector that is furthest from all previous indexed nodes at any point of time throughout the indexing process. To hold the final index, we initialize a list $\mathcal{I}$ that stores the list of links $\mathcal{I}_v$ for each vector $v$ in the dataset $\mathcal{D}$.

\item \textbf{Link Creation}: 
To start the indexing, we choose a random vector from the dataset that will become a node of the graph $\mathcal{G}$. 
The heap $\mathcal{H}$ tracks the nearest known distance $\mathcal{C}_v$, for all unindexed vectors. The top of the heap $\mathcal{H}$ is the vector that is the furthest from all indexed nodes. This vector is known as the create vector and will become the next node in the indexing process. The distance from the create vector to the nearest node is known as the create distance. When a vector becomes a node, it is removed from the  heap $\mathcal{H}$, and we also copy all the links from the local heap $\mathcal{H}_v$ of the vector $v$ and add them into the list $\mathcal{I}_v$. 
The stored links are long at the beginning of indexing, get shorter as indexing progresses, and provide a multiply connected network of links at all length scales for moving around the dataset in descend stage of the \DenseLinkSearch{}.

During the search, this network of links makes it possible to descend from the root node to a node close to the query vector in $\mathcal{O}(log ~\mathcal{N})$ steps. These network links are called \textit{descend links}.

For a vector $v$, the near distance is defined by the link at the top of the local heap $\mathcal{H}_v$, which is the distance to the furthest of the $\mathcal{K}_{index}$ nearest neighbors linked so far. If the heap is not yet full, the near distance is infinite. A vector $\mathcal{A}$ considers another vector $\mathcal{B}$ near if the vector $\mathcal{B}$ is within the vector $\mathcal{A}$'s near distance $\mathcal{R_A}$ (also known as neighborhood radius). Since vectors have different neighborhood radii, it is often the case that a vector $\mathcal{A}$ considers a vector $\mathcal{B}$ near, but the vector $\mathcal{B}$ does not consider the vector $\mathcal{A}$ near.
For the first $\mathcal{K}_{index}$ nodes indexed, no vector's heaps are full, and every vector considers every other vector near. Further indexing adds links to vector heaps, pushing out the longest and therefore  shrinking the near distance.  When neighborhoods shrink to the point that both ends of a link no longer consider the other end near, the link is dropped from the vectors’ heaps and lists. 

Further, to create the links for the new node, we search for the potential neighbors that are not yet nodes, \textit{i.e.}, not yet indexed. A link between vectors $\mathcal{A}$ \& $\mathcal{B}$ will only be created if one or both of the vectors consider the other one near.  The new link is added to heap $\mathcal{H_A}$ of vector $\mathcal{A}$ if vector $\mathcal{A}$  considers vector $\mathcal{B}$ near, \textit{i.e.}, the distance $d_{\mathcal{A}, \mathcal{B}} < \mathcal{R_A}$, otherwise to  list $\mathcal{L_A}$ of vector $\mathcal{A}$.  Likewise, it is added to heap $\mathcal{H_B}$ of vector $\mathcal{B}$ if vector $\mathcal{B}$ considers vector $\mathcal{A}$ near, \textit{i.e.}, the distance $d_{\mathcal{A}, \mathcal{B}} < \mathcal{R_B}$, otherwise to  list $\mathcal{L_B}$ of vector $\mathcal{B}$. Adding a link to a full heap will push the top vector out, shrinking the near distance, $\mathcal{R_A}$ The link that is pushed out is moved to the list if the other endpoint vector still considers this one near. If neither endpoint considers the other near, the link is dropped.

When a vector becomes a node, all $\mathcal{K}_{index}$ of its existing descend links are to existing nodes. Due to the node creation order enforced by the global heap $\mathcal{H}$, the early nodes will have long existing links to far away nodes. Creation always chooses the vector that is the furthest away from all existing nodes so that create distance will continually shrink. Therefore, the nodes created later in the indexing process will have mid-range links to nearer nodes. 
And the last nodes created will have short links to nearest neighbors. 
\item \textbf{Post-processing }: 
At the end of indexing, all nodes have $\mathcal{K}_{index}$ links to their nearest neighbors. These nearest neighbor links are also stored, providing a dense mesh of short links between nearest neighbors. These mesh links make it possible to spread out from a close node to a given query vector and find all the query's nearest neighbors. These mesh links are called \textit{spread links}. The same link may be both a descend and spread link, and no real distinction is made during the search. At the end of indexing, all links are unique and sorted by length to optimize the search.

\end{enumerate}
We have illustrated the indexing algorithm with a running example in Fig. \ref{fig:indexing} and provided the detailed pseudocode for the indexing algorithm in \textbf{Appendix}.

\subsubsection{\DenseLinkSearch{} Algorithm }
Given the dataset $\mathcal{D}$ with $\mathcal{N}$ vectors and its index containing links $\mathcal{I}$, the \DenseLinkSearch{} algorithm finds the $\mathcal{K}_{search}$ nearest neighbors to the query vector $q$ using the following steps:
\begin{figure}[h]
    \centering
    \includegraphics[width=\linewidth]{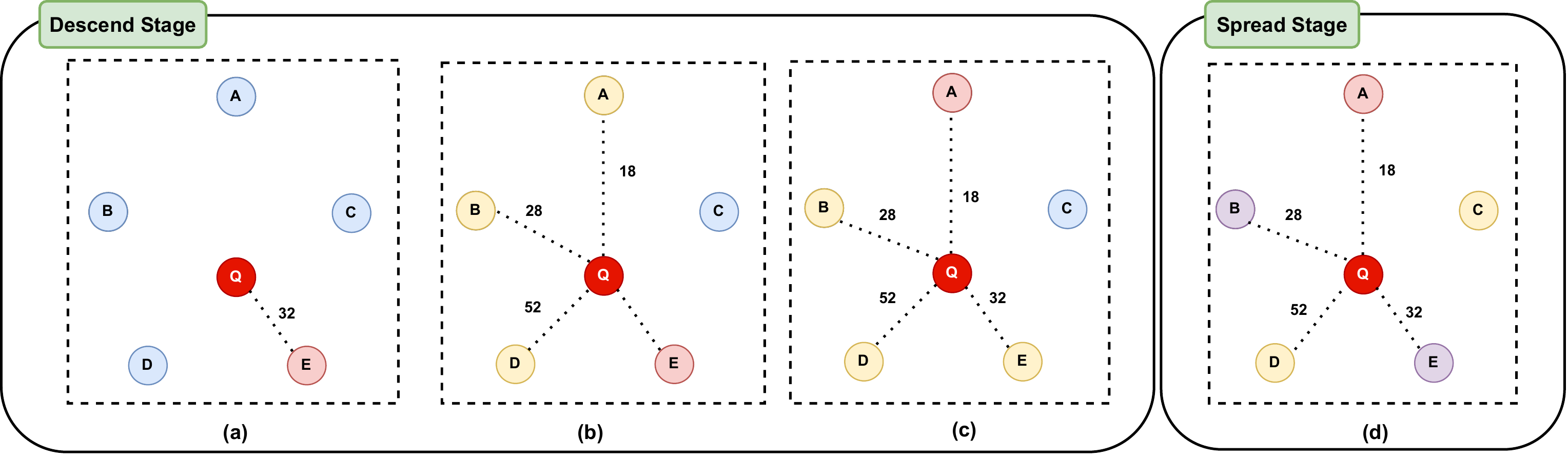}
    \caption{
    The demonstration of \DenseLinkSearch{} that utilizes the index built as shown in Fig. \ref{fig:indexing}. The \DenseLinkSearch{} operates on the Descend and Spread stages to find the $\mathcal{K}_{search}$ nearest neighbors. The input to the search algorithm is a query vector {\tiny \protect\begin{tikz}[baseline=-1mm] \protect\draw[fill=red] circle [radius=0.15] node[font=\sffamily] {\textcolor{white}{Q}}; \protect\end{tikz}} and the pre-built index with the links. In the subfigure (\textbf{a}), the search starts with a random node (here {\tiny \protect\begin{tikz}[baseline=-1mm] \protect\draw[fill=tearose(rose)] circle [radius=0.15] node[font=\sffamily] {\textcolor{black}{E}}; \protect\end{tikz}}) and uses the links of the node {\tiny \protect\begin{tikz}[baseline=-1mm] \protect\draw[fill=tearose(rose)] circle [radius=0.15] node[font=\sffamily] {\textcolor{black}{E}}; \protect\end{tikz}} to compute the distance between the query vector and the node listed in the links of {\tiny \protect\begin{tikz}[baseline=-1mm] \protect\draw[fill=tearose(rose)] circle [radius=0.15] node[font=\sffamily] {\textcolor{black}{E}}; \protect\end{tikz}}. In each step of the Descend stage, the algorithm keeps track of the closest vector to the query. Also, a heap $\mathcal{H}$ of size $\mathcal{K}_{search}$ is maintained to track $\mathcal{K}_{search}$ nearest neighbors throughout the search process. In the subfigure (\textbf{b}), the distance between the query vector and the links ({\tiny \protect\begin{tikz}[baseline=-1mm] \protect\draw[fill=mediumchampagne] circle [radius=0.15] node[font=\sffamily] {\textcolor{black}{A}}; \protect\end{tikz}}, {\tiny \protect\begin{tikz}[baseline=-1mm] \protect\draw[fill=mediumchampagne] circle [radius=0.15] node[font=\sffamily] {\textcolor{black}{B}};\protect\end{tikz}}, and  {\tiny \protect\begin{tikz}[baseline=-1mm] \protect\draw[fill=mediumchampagne] circle [radius=0.15] node[font=\sffamily] {\textcolor{black}{D}};\protect\end{tikz}}) of {\tiny \protect\begin{tikz}[baseline=-1mm] \protect\draw[fill=tearose(rose)] circle [radius=0.15] node[font=\sffamily] {\textcolor{black}{E}}; \protect\end{tikz}} are computed, and the closest vector is updated accordingly. The subfigure (\textbf{c}) shows that the closest node {\tiny \protect\begin{tikz}[baseline=-1mm] \protect\draw[fill=mediumchampagne] circle [radius=0.15] node[font=\sffamily] {\textcolor{black}{A}}; \protect\end{tikz}} is chosen for the next stage of the Descend. Since the distances ($23$ and $32$) between the query and the links of the node {\tiny \protect\begin{tikz}[baseline=-1mm] \protect\draw[fill=tearose(rose)] circle [radius=0.15] node[font=\sffamily] {\textcolor{black}{A}}; \protect\end{tikz}} are greater than distance ($18$) between query and node {\tiny \protect\begin{tikz}[baseline=-1mm] \protect\draw[fill=tearose(rose)] circle [radius=0.15] node[font=\sffamily] {\textcolor{black}{A}}; \protect\end{tikz}}, the search  switches to the Spread stage. The Spread stage is demonstrated in subfigure (\textbf{d}), where, we aim to compute the distance between the query and the links ({\tiny \protect\begin{tikz}[baseline=-1mm] \protect\draw[fill=lightmauve] circle [radius=0.15] node[font=\sffamily] {\textcolor{black}{B}}; \protect\end{tikz}} and {\tiny \protect\begin{tikz}[baseline=-1mm] \protect\draw[fill=lightmauve] circle [radius=0.15] node[font=\sffamily] {\textcolor{black}{E}};\protect\end{tikz}}) of node {\tiny \protect\begin{tikz}[baseline=-1mm] \protect\draw[fill=tearose(rose)] circle [radius=0.15] node[font=\sffamily] {\textcolor{black}{A}}; \protect\end{tikz}}, those have already been computed. At the end of the algorithm, the $\mathcal{K}_{search}$ nearest neighbors can be found in the heap $\mathcal{H}$.
    }
    \label{fig:search}
\end{figure}

\begin{enumerate}
    \item \textbf{Initialization}: Firstly, we initialize a global lookup table ${L}$ that stores the vector $v \in \mathcal{D}$ as key and distance $d_{v, q}$ between vector $v$ and query vector $q$ as the value. We also initialize the global heap $\mathcal{H}$ of size $\mathcal{K}_{search}$ to hold $\mathcal{K}_{search}$ nearest neighbors to the query vector $q$. We start the search from the root vector of the dataset $\mathcal{D}$ and compute the distance between the root vector of dataset $\mathcal{D}$ and the query vector. Initially, the root vector is the nearest neighbor to the query; therefore, the vector is pushed into the global heap $\mathcal{H}$ and also recorded into the ${L}$. We also initialize two global variables $\mathcal{V}_C$ and ${D}_C$. The $\mathcal{V}_C$ and ${D}_C$ are used to keep track of the nearest vector and closest distance (to the query vector) found at any point of time during the NNS.   
    \item \textbf{Descend Stage}: The descend stage utilizes the index $\mathcal{I}$ that is built during the indexing process. During the search, we start with the closest vector $\mathcal{V}_C$ and retrieve all the links $\mathcal{L}$ of $\mathcal{V}_C$ from the $\mathcal{I}$. We traverse each link $l \in \mathcal{L}$, and compute the distance between the vector associated with link $l$ and query vector $q$. During the traversal of the link, we find the local nearest vector and nearest distance from the $\mathcal{L}$ and update the global $\mathcal{V}_C$ and ${D}_C$. 
     We repeat the Descend step as long as we keep getting closer to the query (the closest distance keeps shrinking). Then, we switch to the Spread stage.

    \item \textbf{Spread Stage}: In the Spread stage, we traverse the global heap $\mathcal{H}$ to find the closest vector and its distance to the query vector $q$. Specifically, for each vector $v \in \mathcal{H}$, we extracted their links $\mathcal{L}$ from $\mathcal{L}$ and traverse each link $l \in \mathcal{L}$ in search of the closest vector to the query vector $q$. We continue traversing $\mathcal{H}$ until we find the new  closest vector $\mathcal{V}_N$ for which the distance  ${D}_N$ is smaller than the maximum distanced ($\mathcal{D}_L$) vector ($\mathcal{V}_L$) recorded in the $\mathcal{H}$. Once, we find the closest vector $\mathcal{V}_N$ with their distance to query $D_N$, we compare the $D_N$ with the global closest distance $D_C$. In this case, if the $D_N$ is smaller than the $D_C$ then we update the global closest vector $\mathcal{V}_C$ and closet distance $D_C$ and perform the Descend stage. Otherwise, if the the $D_N$ is smaller than the $D_L$ then we perform the Spread stage again.
    
\end{enumerate}

 We have illustrated the \DenseLinkSearch{} algorithm with a running example in Fig. \ref{fig:search} and provided the detailed pseudocode for the \DenseLinkSearch{} algorithm in \textbf{Appendix}.
\subsection{Image Feature Representation}
To examine the effectiveness of the image feature representation, we performed an extensive experiment considering multiple feature extraction models. We also explore the different feature aggregation techniques to assess their significance in image-based retrieval. In this section, we discuss the feature extraction models and feature aggregators in detail.
\subsubsection{Medical Image Feature Extractors}
\begin{enumerate}
    \item \textbf{Deep Residual Network}: Gradient degradation is a key challenge in training deep neural networks. It is the issue of an increase in training error when layers get added to the network. This causes the low accuracy of the neural network model. With the increase in layers of the network, the gradient computed in the back-propagation step starts to diminish. This problem of vanishing derivatives in deep neural networks is called vanishing gradient descent \citep{hochreiter1998vanishing}. To overcome these problems, \citet{he2016deep}, introduced a deep residual learning framework with deep convolutional neural networks (CNNs) that reformulate the layers as learning residual functions with reference to the layer inputs instead of learning unreferenced functions. 
    Given the input $x$ received from the previous layer, the residual learning framework,  the original mapping is recast into $\mathcal{F}(x)+x$. The formulation of $\mathcal{F}(x)+x$ can be realized by feedforward neural networks with ``shortcut connections''. These connections perform identity mapping, and the outputs of connections are summed to the stacked layers' outputs. In this study, we utilize the pre-trained \resnet{} model as a feature extractor.
    \item \textbf{Vision Transformers}: Inspired by the success of Transformer architecture \citep{vaswani2017attention} in Natural Language Processing (NLP), \citet{dosovitskiy2020image} explore the Transformer architecture with the images and develop Vision Transformer (ViT) . With the Transformer, the images are treated like tokens, as in NLP, by splitting the images into patches. The patch embedding is added with position embedding and passed as input to the Transformer layer. The study conducted by \citet{dosovitskiy2020image} shows that Transformer applied directly to sequences of image patches achieved better results on image classification tasks compared to the CNNs while using fewer computational resources to train the model. We utilize the pre-trained \vitbase{}, \vitlarge{} and \vithuge{} models as image feature extractors.
    \item \textbf{ConvNeXt}: The ConvNeXt \citep{liu2022convnet} is a family of pure CNNs, which is developed by considering the multiple design decisions in Transformers. The ConvNeXt family considers the following key design decisions:
    \begin{enumerate}
        \item The ConvNeXt follows the Swin Transformers \citep{liu2021swin} stage compute ratio strategy. To adopt the strategy, it adjusts the number of blocks sampled in each stage of the network from ($3, 4, 6, 3$) in \resnet{} to ($3, 3, 9, 3$) in ConvNeXt.
         \item The ConvNeXt adopted the grouped convolution approach of ResNeXt \citep{xie2017aggregated} and uses depth-wise convolution, a special case of grouped convolution where the number of groups equals the number of channels.
    \item The ConvNeXt considers the larger kernel-size convolution operations. Varying the kernel sizes from $3$ to $11$, they found the optimal kernel size of $7 \times 7$.
        \item Additionally, ConvNeXt replaces the ReLU activation with the fewer GELU \citep{hendrycks2016gaussian} activation functions and fewer batch normalization \citep{santurkar2018does} layers.
    \end{enumerate}
\end{enumerate}

In this work, we utilized pre-trained ConvNeXt-B, ConvNeXt-L, and ConvNeXt-XL as image feature extractors.
\subsubsection{Feature Aggregations} \label{sec:feature-aggregators}
Given an image $\mathcal{M}$  and an image feature extractor $\mathcal{G}(\mathcal{M}; \theta)$, where $\theta$ denotes the feature extractor parameters that are frozen during feature extraction, we first pre-process the image  and transform the image $\mathcal{M}$ into $\mathcal{X} \in \mathcal{R}^{c \times w \times h}$, where $c$, $w$ and $h$ are number of channels, width, and height of the image respectively. The feature extractor\footnote{We are focused here on CNNs-based feature extractors.}  $\mathcal{G}$ takes $\mathcal{X}$ as input and produces the 3D tensor $x \in \mathcal{R}^{K \times W \times H}$, where $K$ is the number of feature maps in the last layer of feature extractor, $W$ and $H$ refers to the width and height of the feature map respectively.

\begin{enumerate}
    \item \textbf{Max Pooling}: In max pooling, the maximum value of the spatial feature activation for each feature map is considered for feature representation. Formally,
\begin{equation}
f^{(max)} = [f_1^{(max)}, f_2^{(max)}, \ldots f_K^{(max)}]^\top
\label{equ:max_pooling}
\end{equation}
where $f_k^{(max)} = max_{y \in x_{k}}(y)$ is the value obtained after applying $max$ operation on $k^{th}$ feature map $x_k \in \mathcal{R}^{W \times H}$.
    \item \textbf{Sum Pooling}: The sum pooling aims to obtain the feature representation by summing the value of the spatial feature activation for each. Formally,
\begin{equation}
f^{(sum)} = [f_1^{(sum)}, f_2^{(sum)}, \ldots f_K^{(sum)}]^\top
\label{equ:sum_pooling}
\end{equation}
where $f_k^{(sum)} = sum_{y \in x_{k}}(y)$ is the value obtained after applying $sum$ operation on $k^{th}$ feature map $x_k \in \mathcal{R}^{W \times H}$.

\item \textbf{Mean Pooling}: In mean pooling, the average value of the spatial feature activation for each feature map is considered for feature representation. Formally,
\begin{equation}
f^{(mean)} = [f_1^{(mean)}, f_2^{(mean)}, \ldots f_K^{(mean)}]^\top
\label{equ:mean_pooling}
\end{equation}
where $f_k^{(mean)} = mean_{y \in x_{k}}(y)$ is the value obtained after applying $mean$ operation on $k^{th}$ feature map $x_k \in \mathcal{R}^{W \times H}$.

 \item \textbf{Generalized Mean Pooling}: It is the generalization of the pooling technique where generalized mean \citep{dollar2009integral,radenovic2018fine} is used to derive the pooling function. Formally,
 \begin{align}
f^{(gmean)} &= [f_1^{(gmean)}, f_2^{(gmean)}, \ldots f_K^{(gmean)}]^\top \\
\text{where}  \quad f_k^{(gmean)} &= \left( \frac{1}{|x_{k}|}\sum_{y \in x_{k}}\!y^{p_k} \right)^\frac{1}{p_k}
\label{equ:gmean_pooling}
\end{align}
where $p_k$ is a hyper-parameter.
\item \textbf{Spatial-wise Attention}: In spatial-wise attention, we model the importance of each spatial feature by computing the weight. The final feature is generated by considering the spatial weight. We build an attention matrix $\alpha \in \mathcal{R}^{W \times H}$. The sum of any row or column of the matrix $\alpha$ is $1$, which signifies the importance of each spatial position. First, we compute the weight matrix $w \in \mathcal{R}^{W \times H}$, as follows:
\begin{equation}
    w  = \sum_{k=1}^{K} y_k
\end{equation}
Then we apply the \texttt{row-wise softmax} and \texttt{column-wise softmax} operation on $w$ to obtain the attention matrix $\alpha$. The final aggregated features $f^{spatial}$ are obtained as follows:
\begin{equation}
f^{(spatial)} = [f_1^{(spatial)}, f_2^{(spatial)}, \ldots f_K^{(spatial)}]^\top
\label{equ:spatial_pooling}
\end{equation}
where $f_k^{(spatial)} = mean(x_k * \alpha$), obtained after applying element-wise multiplication on $x_k$ and $\alpha$. The $mean$ operation is used to aggregate the feature activation.

\item \textbf{Channel-wise Attention}: In the channel-wise attention, we model the importance of each feature map (channel) by computing the weight. The final feature is generated by considering the channel weight. We build an attention matrix $\beta \in \mathcal{R}^{K}$. The sum of each element of $\beta$ is $1$, which signifies the importance of each feature map. Similar to the spatial-wise attention; first, we compute the weight matrix $w \in \mathcal{R}^{K}$, as follows:
\begin{equation}
    w^{k}  = \sum_{i=1}^{W} \sum_{j=1}^{H} a_{i,j}^{k}
\end{equation}
where $a_{i,j}^{k}$ is an element of $k^{th}$ feature map. Then we apply the \texttt{softmax} operation on $w$ to obtain the attention matrix $\beta$. The final aggregated features $f^{channel}$ are obtained as follows:
\begin{equation}
f^{(channel)} = [f_1^{(channel)}, f_2^{(channel)}, \ldots f_K^{(channel)}]^\top
\label{equ:channel_pooling}
\end{equation}
where $f_k^{(channel)} = mean(x_k \times \beta$), obtained after applying scalar multiplication on $x_k$ with $\beta$. The $mean$ operation is used to aggregate the feature activation.

For the case of the ConvNeXt feature extractor, we performed a detailed investigation on the ConvNeXt layers. We observe that pre-trained LayerNorm \citep{ba2016layer} weights from the last convolution layer can be exploited to obtain better feature representation. Therefore, we pool the ConvNeXt features as follows:
\begin{equation}
\label{equ:connext_pooling}
\begin{split}
f^{(mean)} &= [f_1^{(mean)}, f_2^{(mean)}, \ldots f_K^{(mean)}]^\top \\
f^{(lmean)} &= \text{\texttt{LayerNorm}}(\sigma(f^{(mean)})) \\
\end{split}
\end{equation}
where $f_k^{(mean)} = mean_{y \in x_{k}}(y)$ is the value obtained after applying $mean$ operation on $k^{th}$ feature map $x_k \in \mathcal{R}^{W \times H}$. $\sigma$ denotes the $sigmoid$ activation function. \texttt{LayerNorm(.)} denotes the \texttt{LayerNorm} operation whose weights are initialized with the pre-trained weights from ConvNeXt model.

\end{enumerate}
\section{Experimental Setup}
This section describes the experimental setups for NNS and medical image feature extractors for image retrieval. 
\subsection{Nearest Neighbor Search}
\subsubsection{Index and \DenseLinkSearch{} Computational Details}
We ran all the experiments on a computing node having $72$ CPU, $36$ cores, and $256$ GB RAM. The \DenseLinkSearch{} has $\mathcal{K}_{index}$ and $\mathcal{K}_{search}$ as two key hyper-parameters. The former denotes the number of nearest neighbors calculated and kept in the index, while the latter represents the number of nearest neighbors found by \DenseLinkSearch{} during the search. We have provided the best hyper-parameters values for each NNS approaches in Tables \ref{tab:nns-best-hyper-parameters-1}, and \ref{tab:nns-best-hyper-parameters-2} in the \textbf{Appendix}.
\subsubsection{Datasets for NNS}  \label{sec:nns-dataset}
We evaluated our NNS approach on multiple benchmark datasets from UCI repository\footnote{\url{https://archive.ics.uci.edu/ml/index.php}} and from the  ANN-Benchmark\footnote{\url{https://github.com/erikbern/ann-benchmarks}}. The datasets are diverse in the number of samples (minimum 10K, maximum 12.85M) as well as the dimensions of the sample (minimum 20, maximum 1536). The datasets MNIST, FMNIST, SIFT, and GIST come with the train and test split; we split the remaining datasets into train and test. We use the training datasets to build the indexes and the test datasets to find the $10$ nearest neighbors for each query. We have provided the details of each dataset along with their size and dimensions in Table \ref{tab:NNS-data}.
\paragraph{OpenI Datasets} The OpenI datasets comprises images from the Open Access (OA) subset of PubMed Central\footnote{\url{https://www.ncbi.nlm.nih.gov/pmc/tools/openftlist/}} (PMC). We have curated around $4,222,779$ million articles from PMC OA and extracted $14,908,095$ images from the curated articles. The scientific articles also contain multi-panel images. The multi-panel images are further split into single-panel images using the panel segmentation approach \citep{demner2012design} resulting in a total of $24,625,466$ images. These images also contain graphical illustrations such as charts and graphs. We employ our in-house modality detector \citet{rahman2013multimodal} that categorizes each image into one of the eight modality classes \citep{rahman2013multimodal}. We excluded the images that have been predicted as graphical and only considered medical non-graphical images. This process yields  $12,851,263$ images. The medical image retrieval task experiments described in Section \ref{section: results-feature-extractors} showed that the ConvNeXt-L (IN-22K) model outperforms the existing medical image feature extractors. Therefore, we utilized   ConvNeXt-L (IN-22K) to generate the image features for the OpenI dataset and called it the OpenI-ConvNeXt dataset. To assess the role of feature dimensions for the OpenI dataset, we also generated the image features from the ResNet50 model having $512$ dimensions; we call this OpenI-ResNet dataset.
\begin{table}[]
\resizebox{\textwidth}{!}{%
\begin{tabular}{l|c|c|c|c}
\hline
\textbf{Dataset} & \textbf{\# Samples} & \textbf{Dimension} & \textbf{\# Build Index} & \textbf{\# Query} \\ \hline \hline
Artificial \citep{botta1993learning}       & 10,000        & 40         & 9,000                      & 1,000                \\ 
Faces  \citep{ucidataset}          & 10,304        & 20         & 9,304                      & 1,000                \\ 
Corel \citep{ortega1998supporting}          & 68,040       & 32         & 58,040                     & 10,000               \\ 
MNIST  \citep{lecun1998gradient}          & 70,000        & 784        & 60,000                     & 10,000              \\ 
FMNIST \citep{xiao2017fashion}           & 70,000         & 784        & 60,000                     & 10,000              \\ 
TinyImages \citep{ucidataset}       & 100,000       & 384        & 90,000                     & 10,000               \\
CovType  \citep{blackard1999comparative}        & 581,012       & 54         & 571,012                    & 10,000               \\ 
Twitter  \citep{ucidataset}        & 583,250       & 78         & 573,250                    & 10,000              \\
YearPred \citep{ucidataset}        & 515,345       & 90         & 505,345                    & 10,000               \\ 

SIFT    \citep{jegou2010product}         & 1,000,000         & 128        & 990,000                    & 10,000                   \\
GIST    \citep{jegou2010product}         & 1,000,000         & 960        & 999,000                    & 1,000                \\ 
OpenI-ResNet   (Ours)        &   12,851,263         &  512        &                     12,841,263    &     10,000    \\
OpenI-ConvNeXt  (Ours)            &   12,851,263         &   1,536        &                     12,841,263    &     10,000   
\\ \hline \hline
\end{tabular}%
}
\caption{The details of the benchmark and our created OpenI datasets used for the nearest neighbor search experiments.}
\label{tab:NNS-data}
\end{table}

\subsubsection{Evaluation Metrics for NNS} 
We evaluate the performance of NNS using the following two metrics:
\begin{enumerate}
    \item \textbf{Recall@k (R@k)}:  It is the ratio of the true nearest neighbors retrieved in the top-k nearest neighbors by the algorithm to the $k$ true nearest neighbors in the test set.
    In this study, we focused on retrieving the 10 nearest neighbors; therefore, the metric is R@10.
    \item \textbf{Average time per query (ATPQ)}: It measures the average time per query taken by NNS approaches to retrieve the 10 nearest neighbors. We report this metric in milliseconds.
\end{enumerate}
\subsubsection{Baseline NNS Methods}
We compare the performance of our proposed \DenseLinkSearch{} (DLS) approach with the following competitive NNS methods.
\begin{enumerate}
   
        \item \textbf{KD Tree} \citep{friedman1977algorithm} and \textbf{Ball Tree} \citep{fukunaga1975branch}: These are the tree-based approach used to find the nearest neighbors. We have discussed these approaches in Section \ref{section:related-work}. We use the \texttt{scikit-learn} \citep{scikit-learn} implementation\footnote{\url{https://scikit-learn.org/stable/modules/classes.html\#module-sklearn.neighbors}} of KD and Ball Tree.
    \item \textbf{RP Forest} \citep{yan2019k}:
It works on the concepts of random projection tree \citep{dasgupta2008random} where nearest neighbors are found by combining multiple trees with each constructed recursively through a series of random projections. We utilized the \textit{rpForest} implementation\footnote{\url{https://github.com/lyst/rpforest}}. 

 \item \textbf{Facebook Artificial Intelligence Similarity Search (FAISS)} : 
 Faiss \citep{johnson2019billion} is an approximate nearest neighbors implementation of Locality Sensitive Hashing (LSH) \citep{datar2004locality} based indexing, Inverse Vector File (IVF) \citep{babenko2014inverted}, and Product Quantization \citep{jegou2010product}. We use the FAISS implementation\footnote{\url{https://github.com/facebookresearch/faiss}}. 
  \item \textbf{Annoy} \citep{annoy}: We utilized the approximate nearest neighbors method Annoy\footnote{\url{https://github.com/spotify/annoy}} that strives to minimize memory usage. 
  \item \textbf{Multiple Random Projection Trees (MRPT)} \citep{Hyvonen2016}: In MRPT, multiple random projection trees are combined by a voting scheme. The overall idea is to exploit the redundancy in a large number of candidate data points and eventually reduce the number of expensive exact distance computations using independently generated random projections. We utilized the official implementation\footnote{\url{https://github.com/vioshyvo/mrpt}}.  
  
\item \textbf{Hierarchical Navigable Small World (HNSW)} \citep{malkov2018efficient}: This is a graph-based approximate nearest neighbor approach. HNSW builds a multi-layer structure consisting of a hierarchical set of proximity graphs for nested subsets of the stored elements. To obtain the results from HNSW, we utilized the official implementation\footnote{\url{https://github.com/nmslib/hnswlib}}.

\item \textbf{Scalable Nearest Neighbors (ScaNN)} \citep{guo2020accelerating}:
It is a quantization based nearest search method that computes the approximate distance between each data point and query vector. We utilized the official implementation\footnote{\url{https://github.com/google-research/google-research/tree/master/scann}}. 
\end{enumerate}

\subsection{Medical Image Feature Extractors for Image Retrieval Evaluation}
\subsubsection{Feature Extractors Details and Hyper-parameters}
To extract the features for the images, we use pre-trained weights from the ResNet, ViT, and ConvNeXt models. For the ResNet, we use pre-trained ResNet50 weights\footnote{\url{https://www.tensorflow.org/api_docs/python/tf/keras/applications/resnet50/ResNet50}}. We extracted the features from the \texttt{conv5\_block1\_2\_conv} layer of the ResNet50 model. This layer returns the output tensor of shape $512 \times 7 \times 7$. For ViT, we experiment with the ViT-Base (dimension 768), ViT-Large (dimension 1024), and ViT-Huge (dimension 1280) models. Since ViT is a Transformer model, the \texttt{[CLS]} token representation is considered as the image feature representation. For the ConvNeXt model, we experiment with the variants of the ConvNeXt models. We mainly experiment with the model variants trained on ImageNet-1K \citep{imagenet15russakovsky} dataset and pre-trained on ImageNet-22K \citep{ridnik2021imagenet} dataset. We experiment with ConvNeXt-B, ConvNeXt-L and ConvNeXt-XL models that output feature map shapes $1024 \times 9 \times 9$, $1536 \times 9 \times 9$ and $2048 \times 9 \times 9$ respectively. We utilized pre-trained weights of ViT and ConvNeXt models from \texttt{timm} \citep{rw2019timm}. For generalized mean pooling, we set the $p_k$ hyper-parameter value as $2$.
\subsubsection{Image Retrieval Dataset and evaluation} We used the ImageCLEF 2011 dataset \citep{kalpathy2011overview}.The dataset consists of $231,000$ images from PubMed Central articles, $30$ textual and visual queries, and relevance judgments. For the visual queries, 2-3 sample images (enumerated) are provided. In this study, we  used only the first image as visual query to retrieve the relevant images from the pool of $231,000$ images. For each image, we extract the features from the pre-trained models and rank the images based on the cosine similarity between the query image and database image. 
As in ImageCLEF evaluations, we evaluated the performance using Mean Average Precision, Precision@k, R-precision, and binary preference. The metrics are defined as follows:
\begin{enumerate}
\item \textbf{Precision at k (P@k): } 
Precision at $k$ is the proportion of system retrieved images that are correct.
\begin{equation}
 \text{P@k}=\frac{\text
{\# of correctly retrieved images
in the top-k}}{\text{\# of retrieved images}}
\end{equation}
    \item  \textbf{Mean Average Precision (MAP): } MAP is the average precision averaged across a set of queries.
\begin{equation}
    \text{MAP}=\frac{1}{|Q|} \sum_{j=1}^{|Q|} \frac{1}{m_j} \sum_{k=1}^{m_{j}} P@k_{j}
\end{equation}
where $Q$ is the total set of queries, $m_j$ is the number of correct images returned for the $j^{th}$ query.
\item \textbf{R-precision (Rprec) \citep{buckley2004retrieval}:}  It is deﬁned as the precision of the retrieval system after $R$ documents are retrieved where R is the number of relevant images for the given image query. 
\item \textbf{Binary Preference (bpref) \citep{buckley2004retrieval}:} It computes a preference relation of whether judged relevant images are retrieved ahead of judged irrelevant images. It is defined as follows:
\begin{equation}
     \text{bpref} = \frac{1}{R} \sum_{r}(1- \frac{|n ~ \text{ranked higher than} ~ r|}{R})
\end{equation}
Where $N$ and $R$ are the numbers of judged non-relevant and relevant images, respectively. The notation $r$ is a relevant image, and $n$ is a member of the ﬁrst $R$ judged non-relevant images as retrieved by the system.
We use the \texttt{trec\_eval} evaluation tool\footnote{\url{https://trec.nist.gov/trec\_eval/}} to compute the aforementioned metrics. 
\end{enumerate}
\section{Results and Discussion}

\subsection{Nearest Neighbor Search}
The detailed results comparing the proposed DLS with tree-based and approximate NNS methods are shown in Tables \ref{tab:nns-tree-results} and \ref{nns_main_results} respectively. The best-performing FAISS approach on the respective dataset is shown in Table \ref{nns_main_results}, and the performance of the multiple FAISS approaches are shown in Table \ref{tab:nns-faiss-results}. We have highlighted the FAISS family approach that yield the best performance on the respective dataset in Table \ref{tab:nns-faiss-results}. To report the results (ATPQ and R@10) using the DLS approach on each dataset, we ran the experiments three times and reported the mean value of the results.

For the Faces dataset, our approach outperformed the most competitive approaches (ScaNN and MRPT) with an Avg value of $0.10$ and an R@10 value of $99.20$. The tree-based approaches also reported the R@10 $\geq 99\%$; however, their ATPQ was very high ($0.216$ for KD Tree and $0.214$ for the Ball Tree). Similarly, for the datasets (MNIST, FMNIST, Corel) having size $\approx 70,000$, our approach outperformed the counterpart NNS approaches. We observe that approximate NNS approaches MRPT and HNSW also reported the competitive R@10 values ($\geq 99\%$) however, the proposed DLS approach has lower ATPQ with $\geq 99\%$ R@10 values on MNIST, FMNIST, and Corel datasets. On the TinyImages (size=$100,000$ and dimension=$384$), our approach reported the ATPQ of $3.7$ ms with an R@10 of $99.10\%$; the closet competitive approach ScaNN reported the ATPQ of $0.649$ ms; however, their R@10 was $98.6\%$. The other competitive approach MRPT obtained the R@10 of $99.99\%$ on the TinyImages dataset; however, their ATPQ was $14.31$ ms.

\begin{table}[]
\resizebox{\textwidth}{!}{%
\begin{tabular}{l|cc|cc|cc|cc|cc}
\hline
\multirow{3}{*}{\diagbox[width=8em, height=5.3em]{\textbf{Dataset}}{\textbf{Method}}} & \multicolumn{2}{c|}{\textbf{Brute Search}}                                                                                 & \multicolumn{2}{c|}{\textbf{KD Tree}}                                                                                      & \multicolumn{2}{c|}{\textbf{Ball Tree}}                                                                                    & \multicolumn{2}{c|}{\textbf{RP Forest}}                                                                                    & \multicolumn{2}{c}{\textbf{DLS}}                                                                     \\ \cline{2-11} 
\multicolumn{1}{c|}{}                                 & \multicolumn{1}{c|}{\textbf{\begin{tabular}[c]{@{}c@{}}ATPQ\\ (ms) 	$\downarrow$ \end{tabular}}} & \multicolumn{1}{l|}{\textbf{\begin{tabular}[c]{@{}c@{}}R@10\\ (\%) 	$\uparrow$ \end{tabular}}} & \multicolumn{1}{c|}{\textbf{\begin{tabular}[c]{@{}c@{}}ATPQ\\ (ms) 	$\downarrow$ \end{tabular}}} & \multicolumn{1}{l|}{\textbf{\begin{tabular}[c]{@{}c@{}}R@10\\ (\%) 	$\uparrow$ \end{tabular}}}  & \multicolumn{1}{c|}{\textbf{\begin{tabular}[c]{@{}c@{}}ATPQ\\ (ms) 	$\downarrow$ \end{tabular}}} & \multicolumn{1}{l|}{\textbf{\begin{tabular}[c]{@{}c@{}}R@10\\ (\%) 	$\uparrow$ \end{tabular}}} & \multicolumn{1}{c|}{\textbf{\begin{tabular}[c]{@{}c@{}}ATPQ\\ (ms) 	$\downarrow$ \end{tabular}}} & \multicolumn{1}{l|}{\textbf{\begin{tabular}[c]{@{}c@{}}R@10\\ (\%) 	$\uparrow$ \end{tabular}}} & \multicolumn{1}{c|}{\textbf{\begin{tabular}[c]{@{}c@{}}ATPQ\\ (ms) 	$\downarrow$ \end{tabular}}} & {\textbf{\begin{tabular}[c]{@{}c@{}}R@10\\ (\%) 	$\uparrow$ \end{tabular}}} \\ \hline \hline
\textbf{Artificial}                                    & \multicolumn{1}{c|}{0.860}                                                                 &    100                                & \multicolumn{1}{c|}{0.471}                                                            & 100.0                              & \multicolumn{1}{c|}{0.471}                                                            & 100.0                              & \multicolumn{1}{c|}{0.776}                                                            & 47.79                              & \cellcolor{lightgreen}0.460                                                                 &     \cellcolor{lightgreen} 99.30         \\ 
\textbf{Faces}                                         & \multicolumn{1}{c|}{0.560}                                                                 &    100                                & \multicolumn{1}{c|}{0.216}                                                            & 99.98                              & \multicolumn{1}{c|}{0.214}                                                            & 99.98                              & \multicolumn{1}{c|}{0.934}                                                            & 48.13                              &\cellcolor{lightgreen}0.100                                                                &  \cellcolor{lightgreen}99.20             \\ 
\textbf{Corel}                                         & \multicolumn{1}{c|}{4.200}                                                                 &           100                         & \multicolumn{1}{c|}{2.432}                                                            & 99.99                              & \multicolumn{1}{c|}{2.383}                                                            & 99.99                             & \multicolumn{1}{c|}{6.880}                                                            & 68.31                              & \cellcolor{lightgreen}{0.090}                                                                 &   \cellcolor{lightgreen} 99.50           \\ 
\textbf{MNIST}                                         & \multicolumn{1}{c|}{100}                                                                 &            100                        & \multicolumn{1}{c|}{87.30}                                                            & 100.0                              & \multicolumn{1}{c|}{87.24}                                                            & 100.0                              & \multicolumn{1}{c|}{31.48}                                                            & 71.74                              & \cellcolor{lightgreen}0.850                                                               &   \cellcolor{lightgreen}99.00            \\ 
\textbf{FMNIST}                                        & \multicolumn{1}{c|}{100}                                                                 &     100                               & \multicolumn{1}{c|}{88.92}                                                            & 100.0                              & \multicolumn{1}{c|}{88.70}                                                            & 100.0                              & \multicolumn{1}{c|}{38.20}                                                            & 47.17                              & \cellcolor{lightgreen}0.780                                                                &     \cellcolor{lightgreen} 99.30         \\ 
\textbf{CovType}                                       & \multicolumn{1}{c|}{74.00}                                                                 &           100                         & \multicolumn{1}{c|}{46.21}                                                            & 99.99                              & \multicolumn{1}{c|}{46.14}                                                            & 99.99                             & \multicolumn{1}{c|}{40.43}                                                            & 73.93                              & \cellcolor{lightgreen}0.360                                                                 &    \cellcolor{lightgreen}  99.10         \\
\textbf{TinyImages}                                    & \multicolumn{1}{c|}{76.00}                                                                 &              100                      & \multicolumn{1}{c|}{62.93}                                                            & 99.99                              & \multicolumn{1}{c|}{63.01}                                                            & 99.99                             & \multicolumn{1}{c|}{34.16}                                                            & 35.35                              & \cellcolor{lightgreen}{3.700}                                                                 &     \cellcolor{lightgreen}  99.10        \\ 
\textbf{Twitter}                                       & \multicolumn{1}{c|}{110.0}                                                                 &               100                     & \multicolumn{1}{c|}{70.19}                                                            & 99.60                              & \multicolumn{1}{c|}{70.06}                                                            & 99.60                             & \multicolumn{1}{c|}{78.01}                                                            & 18.68                              & \cellcolor{lightgreen}{0.420}                                                                 &   \cellcolor{lightgreen}99.50            \\ 
\textbf{YearPred}                                      & \multicolumn{1}{c|}{120.0}                                                                 &          100                          & \multicolumn{1}{c|}{74.08}                                                            & 99.99                              & \multicolumn{1}{c|}{74.09}                                                            & 99.99                             & \multicolumn{1}{c|}{58.48}                                                            & 25.30                              & \cellcolor{lightgreen}{2.000}                                                                 &     \cellcolor{lightgreen} 99.30         \\ 
\textbf{SIFT}                                          & \multicolumn{1}{c|}{280.0}                                                                 &   100                                 & \multicolumn{1}{c|}{215.4}                                                            & 99.93                              & \multicolumn{1}{c|}{216.0}                                                            & 99.93                             & \multicolumn{1}{c|}{184.5}                                                            & 99.21                              & \cellcolor{lightgreen}{1.600}                                                                 &   \cellcolor{lightgreen}99.70            \\
\textbf{GIST}                                          & \multicolumn{1}{c|}{2200}                                                                 &                 100                   & \multicolumn{1}{c|}{1920}                                                             & 99.91                              & \multicolumn{1}{c|}{1919}                                                             & 99.91                              & \multicolumn{1}{c|}{735.1}                                                            & 35.70                               & \cellcolor{lightgreen}{36.00}                                                                 &     \cellcolor{lightgreen}   99.10       \\
\hline \hline
\end{tabular}%
}
\caption{Comparison of the proposed DLS approach with multiple tree-based nearest neighbors approaches on benchmark datasets. The \textcolor{lightgreen}{highlighted cells} represent the R@10$\geq 99\%$ for which the ATPQ is the lowest amongst all the approaches. Due to the overhead of memory footprint, we could not run the experiments with tree-based approaches  on the OpenI datasets.}
\label{tab:nns-tree-results}
\end{table}

\begin{table}[]
\resizebox{\textwidth}{!}{%
\begin{tabular}{l|ll|cc|cc|cc|cc|cc|cc}
\hline
\multirow{3}{*}{\diagbox[width=11em, height=5.5em]{\textbf{Dataset}}{\textbf{Method}}} &
  \multicolumn{2}{c|}{\textbf{Brute Search}} &
  \multicolumn{2}{c|}{\textbf{FAISS}} &
  \multicolumn{2}{c|}{\textbf{Annoy}} &
  \multicolumn{2}{c|}{\textbf{MRPT}} &
  \multicolumn{2}{c|}{\textbf{HNSW}} &
  \multicolumn{2}{c|}{\textbf{ScaNN}} &
  \multicolumn{2}{c}{\textbf{DLS}} \\ \cline{2-15} 
\multicolumn{1}{c|}{} &
  \multicolumn{1}{c|}{\textbf{\begin{tabular}[c]{@{}c@{}}ATPQ\\ (ms) 	$\downarrow$ \end{tabular}}} &
 {\textbf{\begin{tabular}[c]{@{}c@{}}R@10\\ (\%) 	$\uparrow$ \end{tabular}}} &
  \multicolumn{1}{c|}{\textbf{\begin{tabular}[c]{@{}c@{}}ATPQ\\ (ms) 	$\downarrow$ \end{tabular}}} &
  \multicolumn{1}{l|}{\textbf{\begin{tabular}[c]{@{}c@{}}R@10\\ (\%) 	$\uparrow$ \end{tabular}}} &
  \multicolumn{1}{c|}{\textbf{\begin{tabular}[c]{@{}c@{}}ATPQ\\ (ms) 	$\downarrow$ \end{tabular}}} &
  \multicolumn{1}{l|}{\textbf{\begin{tabular}[c]{@{}c@{}}R@10\\ (\%) 	$\uparrow$ \end{tabular}}} &
  \multicolumn{1}{c|}{\textbf{\begin{tabular}[c]{@{}c@{}}ATPQ\\ (ms) 	$\downarrow$ \end{tabular}}} &
  \multicolumn{1}{l|}{\textbf{\begin{tabular}[c]{@{}c@{}}R@10\\ (\%) 	$\uparrow$ \end{tabular}}} &
  \multicolumn{1}{c|}{\textbf{\begin{tabular}[c]{@{}c@{}}ATPQ\\ (ms) 	$\downarrow$ \end{tabular}}} &
  \multicolumn{1}{l|}{\textbf{\begin{tabular}[c]{@{}c@{}}R@10\\ (\%) 	$\uparrow$ \end{tabular}}} &
  \multicolumn{1}{c|}{\textbf{\begin{tabular}[c]{@{}c@{}}ATPQ\\ (ms) 	$\downarrow$ \end{tabular}}} &
  \multicolumn{1}{l|}{\textbf{\begin{tabular}[c]{@{}c@{}}R@10\\ (\%) 	$\uparrow$ \end{tabular}}} &
  \multicolumn{1}{c|}{\textbf{\begin{tabular}[c]{@{}c@{}}ATPQ\\ (ms) 	$\downarrow$ \end{tabular}}} &
  {\textbf{\begin{tabular}[c]{@{}c@{}}R@10\\ (\%) 	$\uparrow$ \end{tabular}}} \\ \hline \hline
\textbf{Artificial} &
  \multicolumn{1}{c|}{0.860} &100
   &
  \multicolumn{1}{c|}{3.754} &
  49.63 &
  \multicolumn{1}{c|}{1.034} &
  99.20 & \cellcolor{lightgreen}
  0.123 & \cellcolor{lightgreen}
  100.0 &
  \multicolumn{1}{c|}{0.037} &
  71.39 &
  \multicolumn{1}{c|}{0.033} &
  96.65 &
  \multicolumn{1}{c|}{0.460} &99.30
   \\ 
\textbf{Faces} &
  \multicolumn{1}{c|}{0.560} &100
   &
  \multicolumn{1}{c|}{0.044} &
  89.46 &
  \multicolumn{1}{c|}{0.992} &
  99.41 &
  \multicolumn{1}{c|}{0.111} &
  99.97 &
  \multicolumn{1}{c|}{0.019} &
  86.51 &
  \multicolumn{1}{c|}{0.025} &
  96.66 &
 \cellcolor{lightgreen}0.100 &  \cellcolor{lightgreen} 99.20
   \\ 
\textbf{Corel} &
  \multicolumn{1}{c|}{4.200} & 100
   &
  \multicolumn{1}{c|}{0.217} &
  90.50 &
  \multicolumn{1}{c|}{1.306} &
  99.92 &
  \multicolumn{1}{c|}{0.657} &
  99.99 &
  \multicolumn{1}{c|}{0.024} &
  96.62 &
  \multicolumn{1}{c|}{0.078} &
  57.95 &
   \cellcolor{lightgreen}0.090 &  \cellcolor{lightgreen}99.50
   \\ 
\textbf{MNIST} &
  \multicolumn{1}{c|}{100.0} & 100
   &
  \multicolumn{1}{c|}{4.683} &
  86.02 &
  \multicolumn{1}{c|}{2.235} &
  99.29 &
  \multicolumn{1}{c|}{17.28} &
  100.0 &
  \multicolumn{1}{c|}{0.124} &
  93.42 &
  \multicolumn{1}{c|}{0.878} &
  100.0 &
  \cellcolor{lightgreen}0.850 &  \cellcolor{lightgreen}99.00
   \\ 
\textbf{FMNIST} &
  \multicolumn{1}{c|}{100.0} &100
   &
  \multicolumn{1}{c|}{4.986} &
  92.99 &
  \multicolumn{1}{c|}{2.259} &
  99.06 &
  \multicolumn{1}{c|}{17.61} &
  100.0 &
  \multicolumn{1}{c|}{0.119} &
  94.04 &
  \multicolumn{1}{c|}{0.898} &
  100.0 &
   \cellcolor{lightgreen}0.780 & \cellcolor{lightgreen}99.30
   \\ 
\textbf{CovType} &
  \multicolumn{1}{c|}{74.00} &100
   &
  \multicolumn{1}{c|}{6.532}                                                      & 99.22    &
  \multicolumn{1}{c|}{0.575} &
  99.99 &
  \multicolumn{1}{c|}{15.18} &
  99.99 &
  \multicolumn{1}{c|}{0.024} &
  98.63 &
  \multicolumn{1}{c|}{0.081} &
  17.25 &
  \cellcolor{lightgreen} 0.360 &  \cellcolor{lightgreen} 99.10
   \\ 
\textbf{TinyImages} &
  \multicolumn{1}{c|}{76.00} &100
   &
  \multicolumn{1}{c|}{3.074} &
  73.11 &
  \multicolumn{1}{c|}{3.692} &
  86.27 &
  \multicolumn{1}{c|}{14.31} &
  99.99 &
  \multicolumn{1}{c|}{0.140} &
  63.16 &
  \multicolumn{1}{c|}{0.649} &
  98.96 &
   \cellcolor{lightgreen}3.700& \cellcolor{lightgreen}99.10
   \\ 
\textbf{Twitter} &
  \multicolumn{1}{c|}{110.0} &100
   &
  \multicolumn{1}{c|}{10.19} &
  97.23 &
  \multicolumn{1}{c|}{2.056} &
  98.81 &
  \multicolumn{1}{c|}{20.22} &
  98.95 &
  \multicolumn{1}{c|}{0.051} &
  92.79 &
  \multicolumn{1}{c|}{0.038} &
  2.228 &
 \cellcolor{lightgreen}0.420 &  \cellcolor{lightgreen}99.50
   \\ 
\textbf{YearPred} &
  \multicolumn{1}{c|}{120.0} &100
   &
  \multicolumn{1}{c|}{7.299} &
  90.48 &
  \multicolumn{1}{c|}{1.995} &
  97.28 &
  \multicolumn{1}{c|}{19.50} &
  99.99 &
  \multicolumn{1}{c|}{0.077} &
  79.95 &
  \multicolumn{1}{c|}{0.895} &
  5.818 &
  \cellcolor{lightgreen}2.000 & \cellcolor{lightgreen}99.30
   \\ 
\textbf{SIFT} &
  \multicolumn{1}{c|}{280.0} &100
   &
  \multicolumn{1}{c|}{13.41} &
  89.00 &
  \multicolumn{1}{c|}{2.542} &
  92.57 &
  \multicolumn{1}{c|}{51.61} &
  99.92 &
  \multicolumn{1}{c|}{0.106} &
  79.48 &
  \multicolumn{1}{c|}{3.625} &
  98.72 &
 \cellcolor{lightgreen}1.600 &  \cellcolor{lightgreen}99.70
   \\ 
\textbf{GIST} &
  \multicolumn{1}{c|}{2200} &100
   & 
  \multicolumn{1}{c|}{140.9} &
  78.59 &
  \multicolumn{1}{c|}{9.081} &
  69.19 &
  \multicolumn{1}{c|}{368.9} &
  99.89 &
  \multicolumn{1}{c|}{0.718} &
  54.32 &
  \multicolumn{1}{c|}{26.21} &
  96.20 &
 \cellcolor{lightgreen}36.00 & \cellcolor{lightgreen}99.10
   \\ 
\textbf{OpenI-ResNet} &
  \multicolumn{1}{c|}{19760} &100
   &
 \multicolumn{1}{c|}{869.72}    &                                                   85.80    &
  \multicolumn{1}{c|}{11.96} &
87.03 &
  \multicolumn{1}{c|}{2815.4} &
99.83 &
  \multicolumn{1}{c|}{1.187} &
68.74 &
  \multicolumn{1}{c|}{178.83} &
98.68 &
\cellcolor{lightgreen}18.92 &\cellcolor{lightgreen}99.20 \\
\textbf{OpenI-ConvNeXt} &
  \multicolumn{1}{c|}{3423} &100
   &
  \multicolumn{1}{c|}{2475.7} & 
  90.20  &
  \multicolumn{1}{c|}{18.35} &
82.23  &
  \multicolumn{1}{c|}{7993.4} &
99.48 &
  \multicolumn{1}{c|}{0.7852} &
 54.32 &
  \multicolumn{1}{c|}{530.20} &
99.10 &
\cellcolor{dodgerblue}100.2 & \cellcolor{dodgerblue}97.89
   \\ \hline \hline
\end{tabular}%
}
\caption{Comparison of the proposed DLS approach with approximate nearest neighbors approaches on benchmark datasets. The performance of the best approach from the FAISS family approaches on each dataset is reported under \textbf{FAISS} column. FAISS-IVF performed best amongst all the FAISS's approaches for each dataset except Artificial dataset. Average time per query (ATPQ) is reported in milliseconds, and Recall@10 (R@10) is reported in percentage. The \textcolor{lightgreen}{highlighted cells} represent the R@10$\geq 99\%$ for which the ATPQ is the lowest amongst all the approaches.  The \textcolor{dodgerblue}{highlighted cells} represent the R@10$\geq 97\%$ for which the ATPQ is the lowest amongst all the approaches. Lower ATPQ and higher R@10 are better. }
\label{nns_main_results}
\end{table}

We also analyze the results on the  CovType, Twitter, and YearPred datasets, which have the dataset size $\approx 500,000$. For the CovType, our approach has the ATPQ of $0.36$ ms with R@10 of $99.1\%$. The closest NNS approach HNSW, has the ATPQ of $0.024$ with R@10 of $98.63\%$. Another approximate NNS approach, Annoy, also obtained the R@10 of $99.99\%$ with the ATPQ of $0.575$ ms. Our proposed DLS approach obtained the ATPQ of $0.42$ ms with R@10 of $99.5\%$ on the Twitter dataset. The HNSW recorded the AvgTime of $0.051$ with the R@10 of 92.79. We observe similar patterns on the large-scale datasets (SIFT, GIST, OpenI-ResNet, OpenI-ConvNeXt) with the high dimensions and the dataset sizes in the millions. Our approach outperformed the existing competitive approaches on these high-dimension datasets as well. To summarize, the tree-based approaches consistently produced $\geq 99\%$ R@10; however, the AvgTime were high, which makes them unsuitable for real-time applications, where the latency and accuracy both matter equally. In comparison to the approximate NNS approaches, our proposed DLS approach outperformed them in terms of lower AvgTime and $\geq 99\%$ R@10 on 11 out of 13 datasets.

\subsubsection{Effect of $\mathcal{K}_{index}$ and $\mathcal{K}_{search}$} 
\begin{figure}[h]
\centering
\begin{subfigure}{.5\textwidth}
  \centering
  \includegraphics[width=\linewidth]{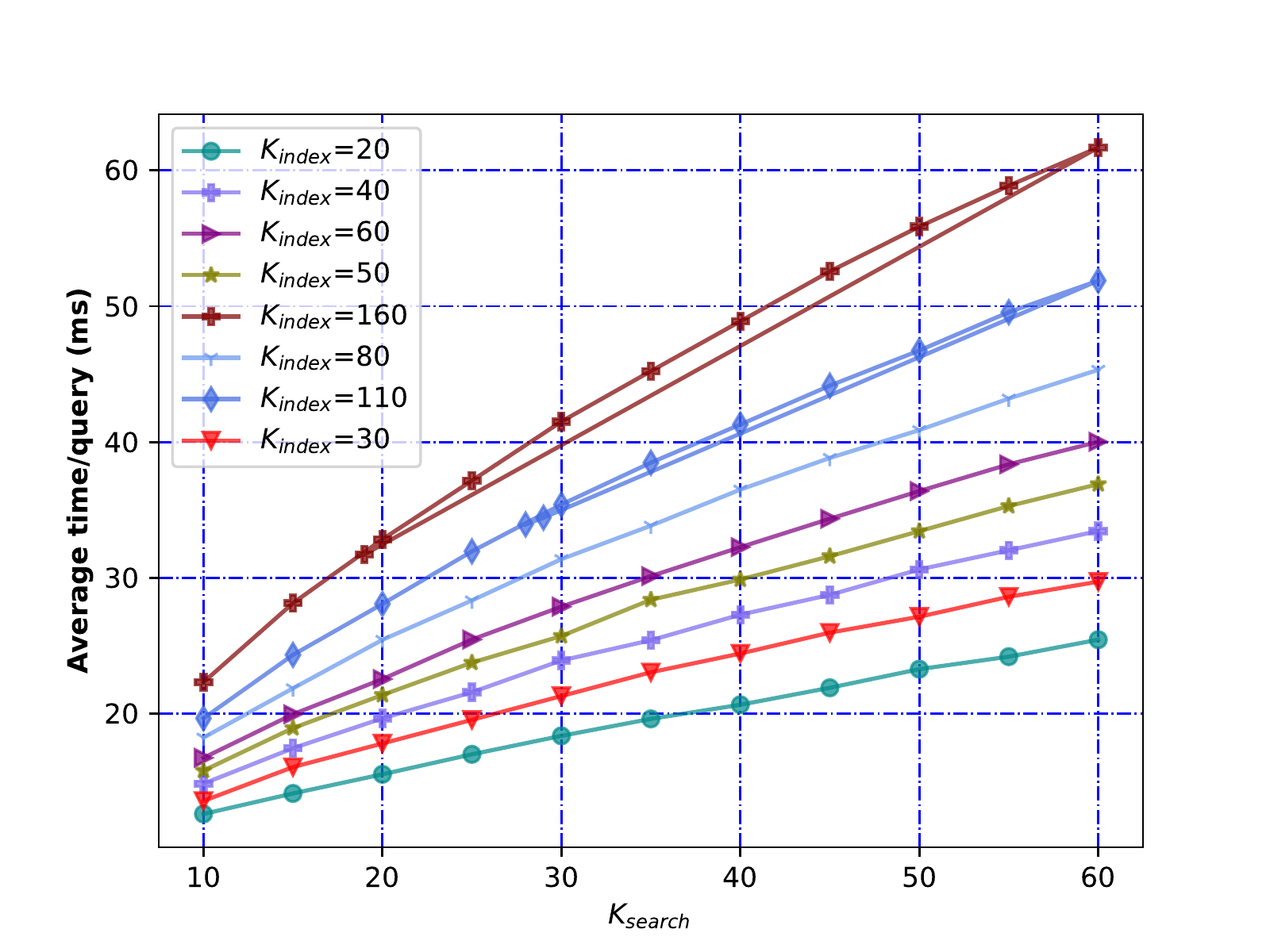}
  \caption{}
  \label{fig:ki_ks_openi_convnext_time}
\end{subfigure}%
\begin{subfigure}{.5\textwidth}
  \centering
  \includegraphics[width=\linewidth]{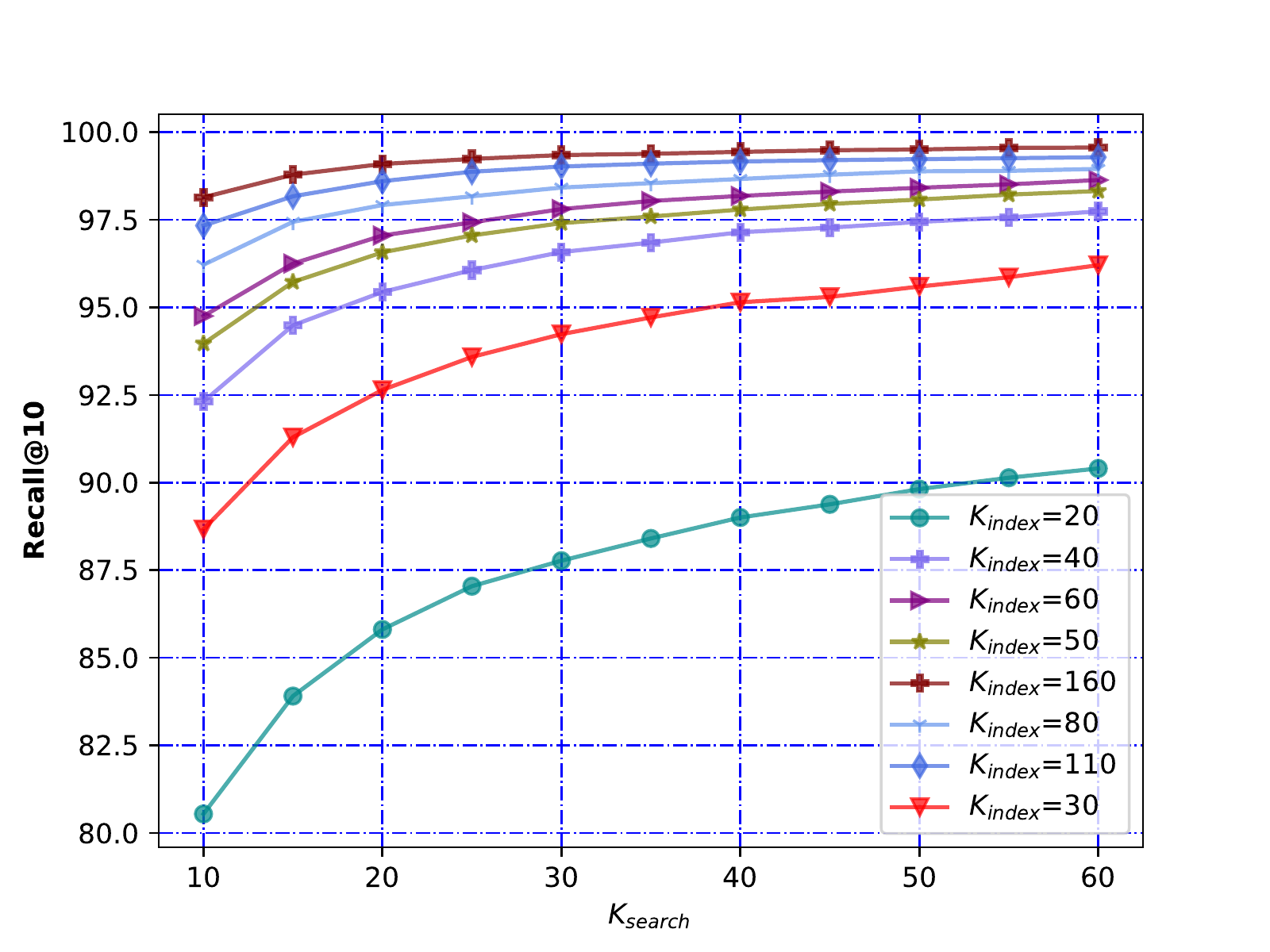}
  \caption{}
  \label{fig:ki_ks_openi_convnext_recall}
\end{subfigure}
\caption{The effect of $\mathcal{K}_{index}$ and $\mathcal{K}_{search}$ on ATPQ (a) and Recall@10 (b) on 1,000,000 samples of OpenI-ConvNeXt dataset.}
\label{fig:ki_ks_openi_convnext}
\end{figure}

\begin{figure}
\centering
\begin{subfigure}{.5\textwidth}
  \centering
  \includegraphics[width=\linewidth]{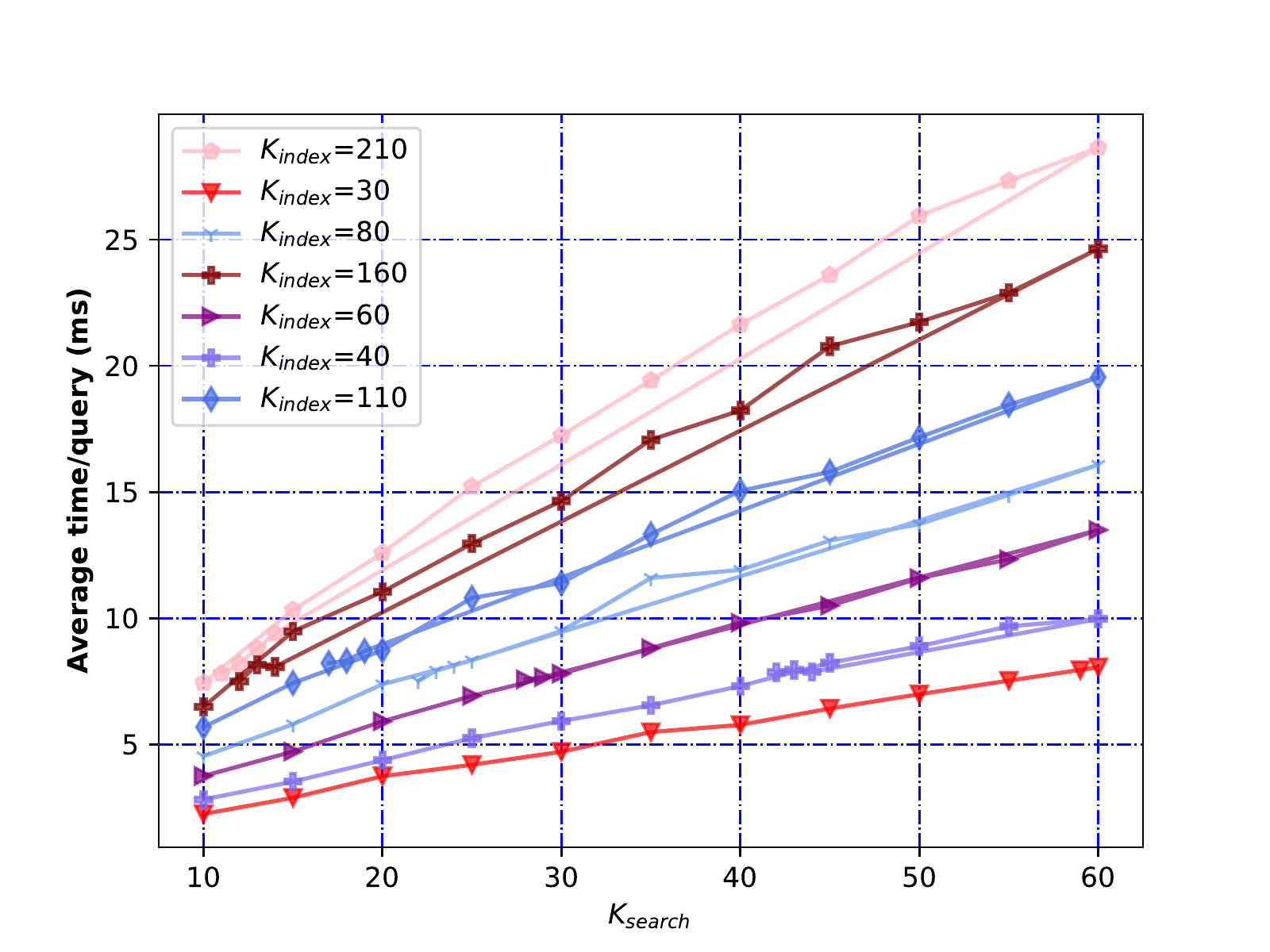}
  \caption{}
  \label{fig:ki_ks_openi_resenet_time}
\end{subfigure}%
\begin{subfigure}{.5\textwidth}
  \centering
  \includegraphics[width=\linewidth]{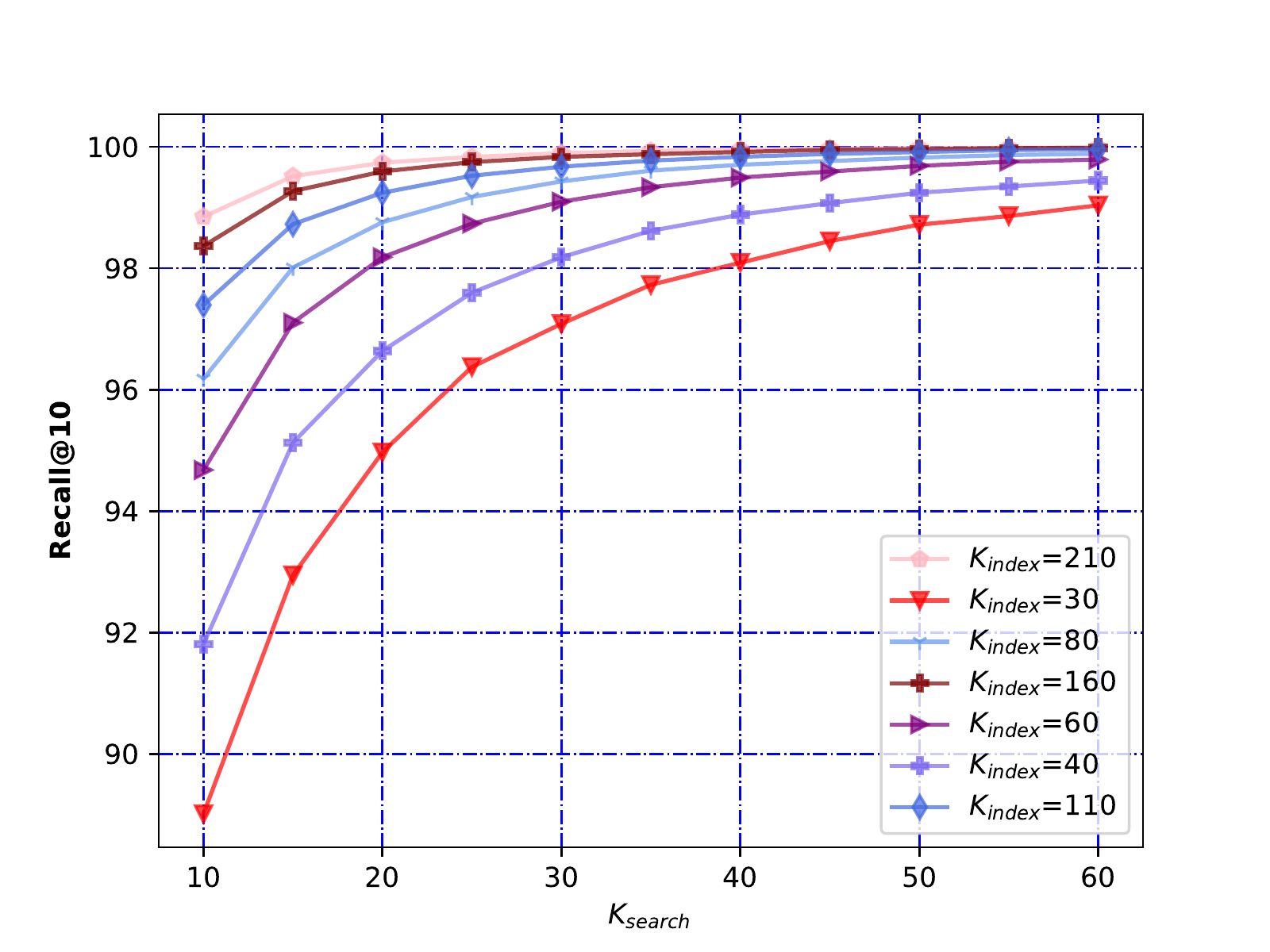}
  \caption{}
  \label{fig:ki_ks_openi_resenet_recall}
\end{subfigure}
\caption{The effect of $\mathcal{K}_{index}$ and $\mathcal{K}_{search}$ on ATPQ (a) and Recall@10 (b) on 1,000,000 samples of the OpenI-ResNet dataset.}
\label{fig:ki_ks_openi_resenet}
\end{figure}

To analyze the effect of the hyper-parameters $\mathcal{K}_{index}$ and $\mathcal{K}_{search}$ of our proposed DLS approach, we performed detailed sensitivity analysis on each benchmark dataset with combination of $\mathcal{K}_{index}$ and $\mathcal{K}_{search}$ values. We observed that search speed is slower when
$\mathcal{K}_{index}$ increases (i.e., includes more nearest neighbor links in the index), and $\mathcal{K}_{search}$ increases (i.e., finds more nearest neighbors with spread during search). In contrast, the R@10 improves as $\mathcal{K}_{index}$ and $\mathcal{K}_{search}$ increase. Similar patterns were observed throughout all the datasets, which confirms that there is a trade-off between search speed and recall; accordingly $\mathcal{K}_{index}$ and $\mathcal{K}_{search}$ should be chosen for the DLS approach as needed. To perform this sensitivity analysis on the OpenI datasets, we chose a subset of the datasets having a size of $1,000,000$ and chose the combinations of $\mathcal{K}_{index}$ and $\mathcal{K}_{search}$ to perform the analysis. We have provided the chart for both the OpenI datasets in Fig. \ref{fig:ki_ks_openi_convnext} and \ref{fig:ki_ks_openi_resenet}.

\subsubsection{Effect of dataset size and feature dimensions }
\begin{figure}[h]
    \centering
    \includegraphics[width=\linewidth]{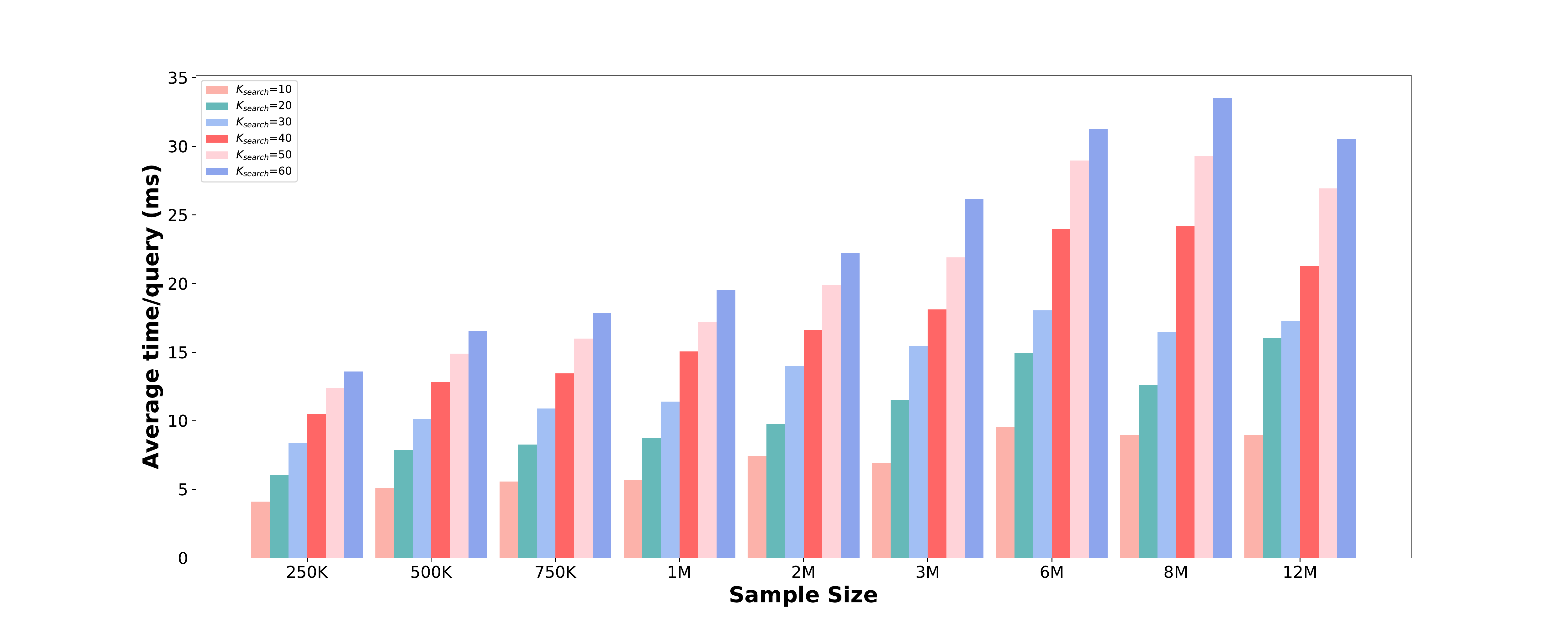}
    \caption{Effect of the sample size on the average time per query for OpenI-ResNet dataset.}
    \label{fig:resnet_sample_size}
\end{figure}

\begin{figure}
    \centering
    \includegraphics[width=\linewidth]{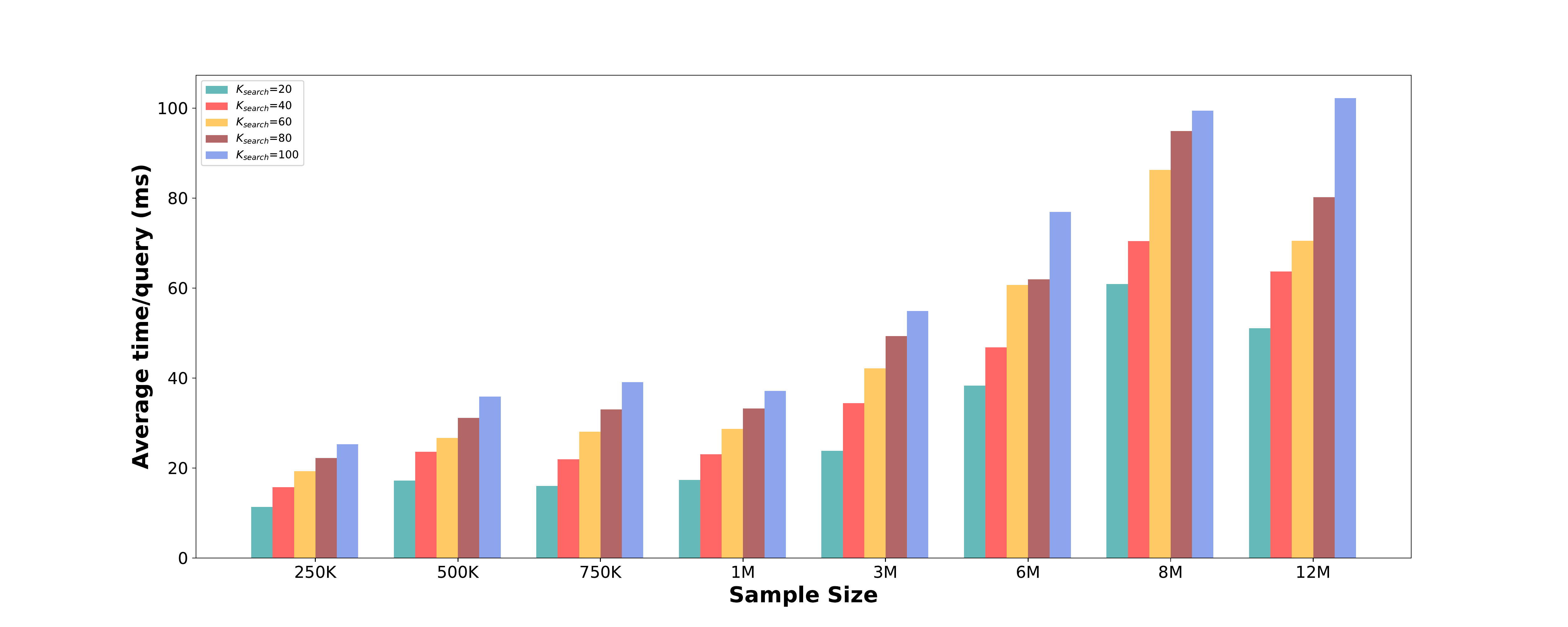}
    \caption{Effect of the sample size on the average time per query for OpenI-ConvNeXt dataset.}
    \label{fig:convnext_sample_size}
\end{figure}

To understand the effect of the dataset size and feature dimensions on the DLS approach, we performed an in-depth analysis by varying the dataset size from $250K$ to $12.85M$ on the OpenI datasets having feature dimensions $512$ and $1,536$. We observe that the average time per query increases with an increase in dataset size as the algorithm needs to perform more distance computations to retrieve the nearest neighbors. With respect to the increase in feature dimensions from $512$ to $1,536$, we notice that the average time per query increases $2.27$, $3.87$, $4.74$ times on $\mathcal{K}_{search}=10$ for $1M$, $3M$ and $12M$ dataset sizes respectively. We have provided the trends for both the OpenI datasets in Fig. \ref{fig:resnet_sample_size} and \ref{fig:convnext_sample_size}.

\subsection{Medical Image Feature Extractors} \label{section: results-feature-extractors}
Table \ref{tab:image_clef_results_models}) presents the results from multiple feature extractors on the ImageCLEF 2011 medical image retrieval task. 

For the ResNet50 model, mean and sum pooling performed better than max pooling on multiple metrics, with a maximum of $0.0518$ and $0.1500$ MAP and P@20 scores using mean pooling with the ResNet50 model. We observed that the ViT-Base model performs better than its counterpart ViT-Large and ViT-Huge models. 
It achieves the maximum MAP of $0.0713$ compared to the ResNet-Mean, which achieved a MAP score of $0.0518$.
The ConvNeXt-L (IN-22K) model achieved the maximum evaluation scores amongst all the variants of ConvNeXt and their counterpart feature extractors. The ConvNeXt-L (IN-22K) model outperformed the best ResNet model by $0.032$, $0.02$, and $0.0674$ in terms of MAP, P@20, and bpref, respectively. It outperformed the best ViT model by $0.0125$, $0.02$, and $0.0126$ in terms of MAP, P@20, and bpref, respectively. Our results are somewhat higher than the best performing system at CLEF 2011 ($0.05$, $0.0533$, and $0.1346$  in terms of MAP, P@10, and bpref, respectively.)

We report the ConvNeXt-L (IN-22K) with mean pooling results (row 1) in Table \ref{tab:image_clef_results_pooling}. We aggregated the features with multiple pooling strategies discussed in Section \ref{sec:feature-aggregators} and report the results in Table \ref{tab:image_clef_results_pooling}. We did not observe any improvements with other feature aggregation strategies over the mean pooling. Since the mean pooling operation performed by the ConvNeXt-L model also utilizes the pre-trained LayerNorm parameters, we also performed the experiments with pre-trained LayerNorm parameters. With this approach, we recorded the improvement in terms of multiple evaluation metrics using the Generalized Mean pooling strategy over mean pooling. The Generalized Mean pooling strategy obtained an improvement of $0.0134$, $0.02$, and $0.01$ in terms of P@5, P@10, and P@20, respectively, in comparison to the mean pooling.

\begin{table}[]
\resizebox{\textwidth}{!}{%
\begin{tabular}{l|cccccc}
\textbf{Models} & \textbf{MAP} & \textbf{P@5} & \textbf{P@10} & \textbf{P@20} & \textbf{Rprec} & \textbf{bpref} \\ \hline  \hline
ResNet with Max-pooling         & 0.0463 & 0.1733 & 0.1533 & 0.1383 & 0.0841 & 0.1455 \\
ResNet with Mean-pooling        & 0.0518 & 0.2000 & 0.1900 & 0.1500 & 0.0838 & 0.1479 \\
ResNet with Sum-pooling        & 0.0517 & 0.2000 & 0.1900 & 0.1500 & 0.0838 & 0.1479 \\ \hline
ViT-Base             & 0.0713 & 0.2133 & 0.1833 & 0.1433 & 0.1109 & 0.2027 \\
ViT-Large            & 0.0679 & 0.1933 & 0.1833 & 0.1500 & 0.0907 & 0.1934 \\
ViT-Huge             & 0.0666 & 0.1600 & 0.1467 & 0.1450 & 0.1146 & 0.1962 \\ \hline
ConvNeXt-B           & 0.0459 & 0.1067 & 0.1033 & 0.0933 & 0.0759 & 0.1595 \\
ConvNeXt-B (IN-22K)  & 0.0699 & 0.2133 & 0.1667 & 0.1500 & 0.1037 & 0.2041 \\
ConvNeXt-L           & 0.0450 & 0.1267 & 0.1133 & 0.0950 & 0.0722 & 0.1568 \\
ConvNeXt-L (IN-22K)  &\cellcolor{lightgreen}0.0838 & \cellcolor{lightgreen}0.2333 & \cellcolor{lightgreen}0.2033 & \cellcolor{lightgreen}0.1700 & \cellcolor{lightgreen}0.1186 & \cellcolor{lightgreen}0.2153 \\
ConvNeXt-XL (IN-22K) & 0.0802 & 0.2067 & 0.1700 & 0.1617 & 0.1091 & 0.2146 \\  \hline
\citet{kalpathy2011overview} & 0.0338 & -- & 0.1500 & -- & -- & 0.0807 \\  \hline
\hline
\end{tabular}%
}
\caption{Performance comparison on CLEF 2011 dataset on medical image retrieval task.  IN-22K refers to the model that is pre-trained on the ImageNet-22K dataset.}
\label{tab:image_clef_results_models}
\end{table}
\begin{table}[]
\resizebox{\textwidth}{!}{%
\begin{tabular}{c|l|cccccc}
\multicolumn{2}{c|}{\textbf{Models}}  & \textbf{MAP} & \textbf{P@5} & \textbf{P@10} & \textbf{P@20} & \textbf{Rprec} & \textbf{bpref} \\ \hline  \hline

\multicolumn{2}{c|}{ConvNeXt-L (IN-22K)} 
  & \cellcolor{lightgreen}0.0838 &0.2333 & 0.2033 &0.1700 &  \cellcolor{lightgreen}0.1186 &  \cellcolor{lightgreen}0.2153 \\ \hline
\multirow{5}{*}{\rotatebox[origin=c]{90}{\begin{tabular}[c]{@{}c@{}}\textbf{Pooling} \end{tabular}}}& 
Max           & 0.0279 & 0.1267 & 0.1067 & 0.0933 & 0.0542 & 0.1290 \\
&Sum           & 0.0698 & 0.2000 & 0.1800 &0.1500 & 0.1033 & 0.1926 \\
&Spatial-wise Attention          & 0.0677 & 0.2133 & 0.1733 & 0.1533 & 0.1024 & 0.1923 \\
&Channel-wise Attention          & 0.0002 & 0.0000 & 0.0000 & 0.0000 & 0.0031 & 0.0143 \\ 
&Generalized Mean         & 0.0521 & 0.1600 & 0.1167 & 0.1117 & 0.0833 & 0.1706 \\  \hline

\multirow{5}{*}{\rotatebox[origin=c]{90}{\begin{tabular}[c]{@{}c@{}}\textbf{LayerNorm} \\ \textbf{with Pooling} \end{tabular}}}& 
Max           & 0.0289 & 0.1133 & 0.1100 & 0.0900 & 0.0614 & 0.1133 \\
&Sum           & 0.0836 & 0.2467 & 0.2133 &0.1833 & 0.1126 & 0.2100 \\
&Spatial-wise Attention          & 0.0821 & 0.2333 & 0.1967 & 0.1783 & 0.1110 & 0.2086 \\
&Channel-wise Attention          &0.0046 & 0.0200 & 0.0233 & 0.0200 & 0.0162 & 0.0554 \\ 
&Generalized Mean         &  \cellcolor{lightgreen}0.0838 &  \cellcolor{lightgreen}0.2467 &  \cellcolor{lightgreen}0.2233 &  \cellcolor{lightgreen}0.1800 & 0.1153 & 0.2129 \\  \hline
\hline
\end{tabular}%
}
\caption{Performance comparison of different feature aggregation (pooling) techniques on CLEF 2011 dataset on medical image retrieval task. The feature maps from model ConvNeXt-L (IN-22K) are extracted, and respective pooling operation was performed to report the numbers. The first row results are reported with mean pooling operation.}
\label{tab:image_clef_results_pooling}
\end{table}

\begin{figure}
\centering
\begin{subfigure}{.48\textwidth}
  \centering
  \includegraphics[width=\linewidth, height=5cm]{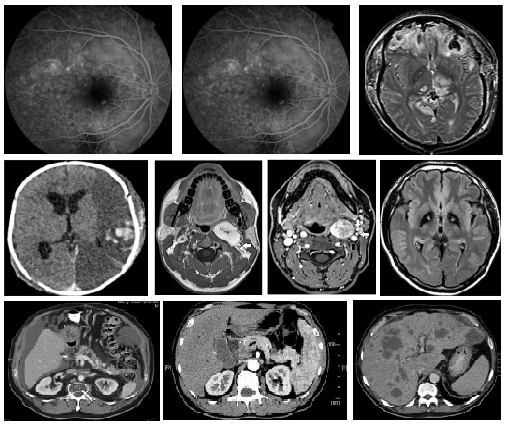}
  \caption{}
  \label{fig:openi-existing}
\end{subfigure}%
\quad
\begin{subfigure}{.48\textwidth}
  \centering
  \includegraphics[width=\linewidth,height=5cm]{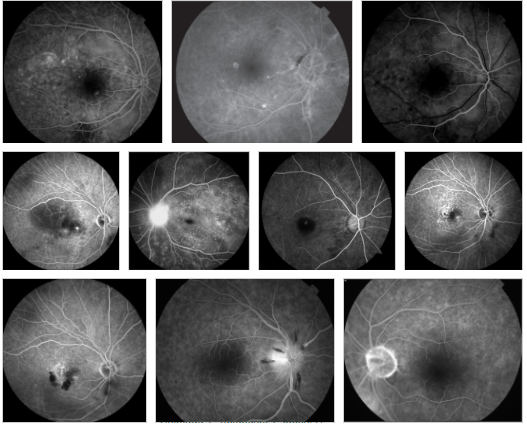}
  \caption{}
  \label{fig:openi-improved}
\end{subfigure}
\caption{Comparison of the nearest images retrieved from the existing OpenI system \textbf{(a)} and our proposed system \textbf{(b)}. The first image in each of the subfigures is the query image, and the remaining are the nearest images in the decreasing order of their similarity to the query image.  }
\label{fig:openi-existing-vs-improved}
\end{figure}

\subsection{Effect of Medical Image Features and \DenseLinkSearch{} on Open-i Service} Open-i service at the National Library of Medicine enables search and retrieval of abstracts and images from the open source literature and biomedical image collections. We aim to improve the Open-i\footnote{\url{https://openi.nlm.nih.gov/}} image search service by the proposed medical feature extraction, and \DenseLinkSearch{} approaches. We have shown the effect of the proposed approaches for retrieving similar images in  Fig. \ref{fig:openi-existing-vs-improved}. It is clearly illustrated that the proposed medical features extraction with the effective \DenseLinkSearch{} is able to retrieve similar \textit{retinal images} given an image of retina as query, while the current \OpenI{}{} system retrieves mostly \textit{brain images} as the nearest images to the same \textit{retinal image}. 

\section{Conclusion}
In this paper, we introduced an effective approach to nearest neighbor search that focuses on traversing the links created during indexing to reduce the distance computation for NNS. We evaluated multiple state-of-the-art approximate NNS algorithms and tree-based algorithms on multiple benchmark datasets in a comprehensive manner. The proposed \DenseLinkSearch{} outperformed the existing approaches, where our approach achieves $\geq 99\%$ R@10 with the lowest average time/query on most of the benchmark datasets. In addition, we explored the role of image feature extractors for medical image retrieval tasks. We experimented with multiple pre-trained vision models and devised an effective approach for image feature extractors that outperformed the pre-trained vision model-based image feature extractors with a fair margin on multiple evaluation metrics.

\section*{Acknowledgements}
This work was supported by the intramural research program at the U.S. National Library of Medicine, National Institutes of Health (NIH), and utilized the computational resources of the NIH HPC Biowulf cluster (\url{http://hpc.nih.gov}). The content is solely the responsibility of the authors and does not necessarily represent the official views of the National Institutes of Health.

\bibliographystyle{model5-names}\biboptions{authoryear}
\bibliography{mybibfile}
\clearpage
\appendix
\section{Hyper-parameters Details}
\begin{table}[h]
\resizebox{\textwidth}{!}{%
\begin{tabular}{l|c|c|c|c|c|c}
\hline
\diagbox{\textbf{Method}}{\textbf{Dataset}}              & \textbf{Artificial}                                                                              & \textbf{Faces}                                                                                   & \textbf{Corel}                                                                                   & \textbf{MNIST}                                                                                  & \textbf{FMNIST}                                                                                 & \textbf{CovType}                                                                                   \\ \hline \hline
\textbf{Ball Tree}     & \texttt{leaf\_size=9K}                                                                                    & \texttt{leaf\_size=4500}                                                                                  & \texttt{leaf\_size=58K}                                                                                   & \texttt{leaf\_size=60K}                                                                                  & \texttt{leaf\_size=45K}                                                                                  & \texttt{leaf\_size=580K}                                                                                    \\ \hline
\textbf{KD Tree}       & \texttt{leaf\_size=9K}                                                                                    & \texttt{leaf\_size=6K}                                                                                    & \texttt{leaf\_size=58K}                                                                                   & \texttt{leaf\_size=30K }                                                                                 & \texttt{leaf\_size=45K}                                                                                  & \texttt{leaf\_size=580K}                                                                                    \\ \hline
\textbf{RP Forest}     & \begin{tabular}[c]{@{}c@{}} \texttt{leaf\_size=900}\\ \texttt{n\_trees=10}\end{tabular}                             & \begin{tabular}[c]{@{}c@{}}\texttt{leaf\_size=600}\\ \texttt{n\_trees=15}\end{tabular}                             & \begin{tabular}[c]{@{}c@{}}\texttt{leaf\_size=3800}\\ \texttt{n\_trees=15}\end{tabular}                            & \begin{tabular}[c]{@{}c@{}} \texttt{leaf\_size=6K}\\ \texttt{n\_trees=10}\end{tabular}                             & \begin{tabular}[c]{@{}c@{}} \texttt{leaf\_size=4K}\\ \texttt{n\_trees=15}\end{tabular}                             & \begin{tabular}[c]{@{}c@{}} \texttt{leaf\_size=29K}\\ \texttt{n\_trees=20}\end{tabular}                               \\ \hline
\textbf{FAISS-LSH}     & \texttt{n\_bits=16384}                                                                                    & \texttt{n\_bits=16384 }                                                                                   & \texttt{n\_bits=4096}                                                                                     & \texttt{n\_bits=4096}                                                                                    & \texttt{n\_bits=4096}                                                                                    & \texttt{n\_bits=4096}                                                                                       \\ \hline
\textbf{FAISS-IVF}     &\texttt{ n\_list=5}                                                                                        & \texttt{n\_list=5}                                                                                        & \texttt{n\_list=5}                                                                                        & \texttt{n\_list=5}                                                                                       & \texttt{n\_list=5}                                                                                       &\texttt{ n\_list=5  }                                                                                        \\ \hline
\textbf{FAISS-IVFPQfs} & \texttt{n\_list=5}                                                                                        & \texttt{n\_list=5}                                                                                        & \texttt{n\_list=10  }                                                                                     & \texttt{n\_list=5}                                                                                       & \texttt{n\_list=5}                                                                                       & \texttt{n\_list=100}                                                                                        \\ \hline
\textbf{Annoy}         & \texttt{n\_trees=500}                                                                                     & \texttt{n\_trees=500}                                                                                     & \texttt{n\_trees=500}                                                                                     & \texttt{n\_trees=500 }                                                                                   & \texttt{n\_trees=500 }                                                                                   & \texttt{n\_trees=300}                                                                                       \\ \hline
\textbf{MRPT}          & -                                                                                                & -                                                                                                & -                                                                                                & -                                                                                               & -                                                                                               & -                                                                                                  \\ \hline
\textbf{HNSW}          & \begin{tabular}[c]{@{}c@{}} \texttt{ef\_construction=100}\\ \texttt{M=50}\end{tabular}                              & \begin{tabular}[c]{@{}c@{}} \texttt{ef\_construction=100}\\ \texttt{M=10}\end{tabular}                              & \begin{tabular}[c]{@{}c@{}} \texttt{ef\_construction=100}\\ \texttt{M=50}\end{tabular}                              & \begin{tabular}[c]{@{}c@{}}\texttt{ef\_construction=100}\\ \texttt{M=50}\end{tabular}                             & \begin{tabular}[c]{@{}c@{}}\texttt{ef\_construction=100}\\ \texttt{M=100}\end{tabular}                            & \begin{tabular}[c]{@{}c@{}}\texttt{ef\_construction=100}\\ \texttt{M=20}\end{tabular}                                \\ \hline
\textbf{ScaNN}         & \begin{tabular}[c]{@{}c@{}}\texttt{n\_leaves=1}\\  \texttt{avq\_threshold=0.20}\\  \texttt{dims\_per\_block=2}\end{tabular} & \begin{tabular}[c]{@{}c@{}}\texttt{n\_leaves=1}\\  \texttt{avq\_threshold=0.20}\\  \texttt{dims\_per\_block=2}\end{tabular} & \begin{tabular}[c]{@{}c@{}}\texttt{n\_leaves=1}\\  \texttt{avq\_threshold=0.20}\\  \texttt{dims\_per\_block=2}\end{tabular} & \begin{tabular}[c]{@{}c@{}}\texttt{n\_leaves=1}\\  \texttt{avq\_threshold=0.2}\\  \texttt{dims\_per\_block=2}\end{tabular} & \begin{tabular}[c]{@{}c@{}}\texttt{n\_leaves=1}\\  \texttt{avq\_threshold=0.2}\\  \texttt{dims\_per\_block=2}\end{tabular} & \begin{tabular}[c]{@{}c@{}}\texttt{n\_leaves=5000}\\  \texttt{avq\_threshold=0.2}\\  \texttt{dims\_per\_block=2}\end{tabular} \\ \hline
\textbf{DLS}           &  \begin{tabular}[c]{@{}c@{}}\texttt{$K_{index}$=50}\\ \texttt{$K_{search}$=20}\end{tabular}                                                                                                &         \begin{tabular}[c]{@{}c@{}}\texttt{$K_{index}$=50}\\ \texttt{$K_{search}$=20}\end{tabular}                                                                                          &            \begin{tabular}[c]{@{}c@{}}\texttt{$K_{index}$=50}\\ \texttt{$K_{search}$=10}\end{tabular}                                                                                      &    \begin{tabular}[c]{@{}c@{}}\texttt{$K_{index}$=40}\\ \texttt{$K_{search}$=15}\end{tabular}                                                                                  &      \begin{tabular}[c]{@{}c@{}}\texttt{$K_{index}$=50}\\ \texttt{$K_{search}$=10}\end{tabular}                                                                                             &    \begin{tabular}[c]{@{}c@{}}\texttt{$K_{index}$=40}\\ \texttt{$K_{search}$=50}\end{tabular}                                                                                                 \\ \hline \hline
\end{tabular}%
}
\caption{The hyper-parameters details that resulted in the best performance of the NNS approaches on benchmark datasets.}
\label{tab:nns-best-hyper-parameters-1}
\end{table}

\begin{table}[h]
\resizebox{\textwidth}{!}{%
\begin{tabular}{l|c|c|c|c|c|c|c}
\hline
\diagbox{\textbf{Method}}{\textbf{Dataset}}                 & \textbf{TinyImages}                                                                             & \textbf{Twitter}                                                                                 & \textbf{YearPred}                                                                               & \textbf{SIFT}                                                                                   & \textbf{GIST}                                                                                   & \textbf{OpenI-ResNet} & \textbf{OpenI-ConvNeXt} \\ \hline \hline
\textbf{Ball Tree}     & \texttt{leaf\_size=60K}                                                                                  & \texttt{leaf\_size=570K}                                                                                  & \texttt{leaf\_size=500K }                                                                                & \texttt{leaf\_size=700K}                                                                                 & \texttt{leaf\_size=700K}                                                                                 &         NA     & NA   \\ \hline
\textbf{KD Tree}       & \texttt{leaf\_size=60K}                                                                                  & \texttt{leaf\_size=570K}                                                                                  & \texttt{leaf\_size=500K}                                                                                 & \texttt{leaf\_size=500K}                                                                                 & \texttt{leaf\_size=500K}                                                                                 &          NA  & NA     \\ \hline
\textbf{RP Forest}     & \begin{tabular}[c]{@{}c@{}}\texttt{leaf\_size=6K}\\ \texttt{n\_trees=15}\end{tabular}                             & \begin{tabular}[c]{@{}c@{}}\texttt{leaf\_size=57K}\\ \texttt{n\_trees=10}\end{tabular}                             & \begin{tabular}[c]{@{}c@{}}\texttt{leaf\_size=30K}\\ \texttt{n\_trees=15}\end{tabular}                            & \begin{tabular}[c]{@{}c@{}}\texttt{leaf\_size=5K}\\ \texttt{n\_trees=200}\end{tabular}                            & \begin{tabular}[c]{@{}c@{}}\texttt{leaf\_size=5K}\\
\texttt{n\_trees=200}\end{tabular}                            &          NA & NA      \\ \hline
\textbf{FAISS-LSH}     & \texttt{n\_bits=4096}                                                                                    & \texttt{n\_bits=4096}                                                                                     & \texttt{n\_bits=4096}                                                                                    &\texttt{ n\_bits=4096  }                                                                                  & \texttt{n\_bits=4096}                                                                                    &    \texttt{n\_bits=4096}  &    \texttt{n\_bits=4096}                  \\ \hline
\textbf{FAISS-IVF}     & \texttt{n\_list=5}                                                                                       & \texttt{n\_list=5}                                                                                        & \texttt{n\_list=5 }                                                                                      & \texttt{n\_list=5}                                                                                       & \texttt{n\_list=5}                                                                                       &       \texttt{n\_list=5}     &       \texttt{n\_list=25}         \\ \hline
\textbf{FAISS-IVFPQfs} & \texttt{n\_list=5}                                                                                       & \texttt{n\_list=100}                                                                                      & \texttt{n\_list=5}                                                                                       & \texttt{n\_list=5}                                                                                       & \texttt{n\_list=5}                                                                                       &          \texttt{n\_list=5}    &          \texttt{n\_list=5}     \\ \hline
\textbf{Annoy}         & \texttt{n\_trees=1000}                                                                                   & \texttt{n\_trees=500}                                                                                      & \texttt{n\_trees=500 }                                                                                    &\texttt{ n\_trees=500 }                                                                                    & \texttt{n\_trees=1000}                                                                                    &      \texttt{n\_trees=1000}&      \texttt{n\_trees=500}             \\ \hline
\textbf{MRPT}          & -                                                                                               & -                                                                                                & -                                                                                               & -                                                                                               & -                                                                                               &          -  & -      \\ \hline
\textbf{HNSW}          & \begin{tabular}[c]{@{}c@{}}\texttt{ef\_construction=100}\\ \texttt{M=100}\end{tabular}                            & \begin{tabular}[c]{@{}c@{}}\texttt{ef\_construction=100}\\ \texttt{M=50}\end{tabular}                              & \begin{tabular}[c]{@{}c@{}}\texttt{ef\_construction=100}\\ \texttt{M=50}\end{tabular}                             & \begin{tabular}[c]{@{}c@{}}\texttt{ef\_construction=100}\\ \texttt{M=200}\end{tabular}                            & \begin{tabular}[c]{@{}c@{}}\texttt{ef\_construction=100}\\ \texttt{M=200}\end{tabular}                            &           \begin{tabular}[c]{@{}c@{}}\texttt{ef\_construction=100}\\ \texttt{M=100} \end{tabular}   & \begin{tabular}[c]{@{}c@{}}\texttt{ef\_construction=100}\\ \texttt{M=100} \end{tabular}   \\ \hline
\textbf{ScaNN}         & \begin{tabular}[c]{@{}c@{}}\texttt{n\_leaves=1}\\  \texttt{avq\_threshold=0.2}\\  \texttt{dims\_per\_block=2}\end{tabular} & \begin{tabular}[c]{@{}c@{}}\texttt{n\_leaves=75}\\  \texttt{avq\_threshold=0.2}\\  \texttt{dims\_per\_block=2}\end{tabular} & \begin{tabular}[c]{@{}c@{}}\texttt{n\_leaves=1}\\  \texttt{avq\_threshold=0.2}\\  \texttt{dims\_per\_block=2}\end{tabular} & \begin{tabular}[c]{@{}c@{}}\texttt{n\_leaves=1}\\  \texttt{avq\_threshold=0.2}\\  \texttt{dims\_per\_block=2}\end{tabular} & \begin{tabular}[c]{@{}c@{}}\texttt{n\_leaves=1}\\  \texttt{avq\_threshold=0.2}\\  \texttt{dims\_per\_block=2}\end{tabular} &     \begin{tabular}[c]{@{}c@{}}\texttt{n\_leaves=1}\\  \texttt{avq\_threshold=0.2}\\  \texttt{dims\_per\_block=2}\end{tabular} & \begin{tabular}[c]{@{}c@{}}\texttt{n\_leaves=1}\\  \texttt{avq\_threshold=0.2}\\  \texttt{dims\_per\_block=2}\end{tabular}           \\ \hline
\textbf{DLS}           &  \begin{tabular}[c]{@{}c@{}}\texttt{$K_{index}$=20}\\ \texttt{$K_{search}$=50}\end{tabular}                                                                                                &         \begin{tabular}[c]{@{}c@{}}\texttt{$K_{index}$=30}\\ \texttt{$K_{search}$=20}\end{tabular}                                                                                          &            \begin{tabular}[c]{@{}c@{}}\texttt{$K_{index}$=30}\\ \texttt{$K_{search}$=25}\end{tabular}                                                                                      &    \begin{tabular}[c]{@{}c@{}}\texttt{$K_{index}$=40}\\ \texttt{$K_{search}$=30}\end{tabular}                                                                                  &      \begin{tabular}[c]{@{}c@{}}\texttt{$K_{index}$=40}\\ \texttt{$K_{search}$=55}\end{tabular}                                                                                             &    \begin{tabular}[c]{@{}c@{}}\texttt{$K_{index}$=110}\\ \texttt{$K_{search}$=35}\end{tabular}& \begin{tabular}[c]{@{}c@{}}\texttt{$K_{index}$=30}\\ \texttt{$K_{search}$=110}\end{tabular}               \\ \hline \hline
\end{tabular}%
}
\caption{The hyper-parameters details that resulted in the best performance of the NNS approaches on benchmark datasets.}
\label{tab:nns-best-hyper-parameters-2}
\end{table}
\newpage
\section{Additional Results}
\begin{table}[h]
\resizebox{\textwidth}{!}{%
\begin{tabular}{l|cc|cc|cc}
\hline
\multirow{3}{*}{\diagbox[width=11em, height=5.5em]{\textbf{Dataset}}{\textbf{Method}}} & \multicolumn{2}{c|}{\textbf{FAISS-LSH}}                                                                                    & \multicolumn{2}{c|}{\textbf{FAISS-IVF}}                                                                                    & \multicolumn{2}{c}{\textbf{FAISS-IVFPQfs}}                                                                                \\ \cline{2-7} 
\multicolumn{1}{c|}{}                                 & \multicolumn{1}{c|}{\textbf{\begin{tabular}[c]{@{}c@{}}ATPQ\\ (ms) 	$\downarrow$ \end{tabular}}} & \multicolumn{1}{l|}{\textbf{\begin{tabular}[c]{@{}c@{}}R@10\\ (\%) 	$\uparrow$ \end{tabular}}} & \multicolumn{1}{c|}{\textbf{\begin{tabular}[c]{@{}c@{}}ATPQ\\ (ms) $\downarrow$\end{tabular}}} & \multicolumn{1}{l|}{\textbf{\begin{tabular}[c]{@{}c@{}}R@10\\ (\%) 	$\uparrow$ \end{tabular}}} & \multicolumn{1}{c|}{\textbf{\begin{tabular}[c]{@{}c@{}}ATPQ\\ (ms) $\downarrow$\end{tabular}}} & \multicolumn{1}{l}{\textbf{\begin{tabular}[c]{@{}c@{}}R@10\\ (\%) 	$\uparrow$ \end{tabular}}} \\ \hline \hline
\textbf{Artificial}                                    & \multicolumn{1}{c|}  {\cellcolor{lightgreen}{3.754}}                                                      &  \cellcolor{lightgreen}{49.63}                              & \multicolumn{1}{c|}{0.061}                                                    & 39.86                              & \multicolumn{1}{c|}{0.030}                                                    & 30.11                              \\ 
\textbf{Faces}                                         & \multicolumn{1}{c|}{3.772}                                                      & 47.56                              & \multicolumn{1}{c|}{\cellcolor{lightgreen}{0.044}}                                                    & \cellcolor{lightgreen}{89.46}                              & \multicolumn{1}{c|}{0.027}                                                     & 42.27                              \\ 
\textbf{Corel}                                         & \multicolumn{1}{c|}{6.699}                                                      & 66.23                             & \multicolumn{1}{c|}{\cellcolor{lightgreen}{0.217}}                                                     & \cellcolor{lightgreen}{90.50}                         & \multicolumn{1}{c|}{0.0260}                                                    & 31.22                         \\ 
\textbf{MNIST}                                         & \multicolumn{1}{c|}{7.396}                                                      & 68.16                              & \multicolumn{1}{c|}{\cellcolor{lightgreen}{4.683}}                                                      & \cellcolor{lightgreen}{86.02}                             & \multicolumn{1}{c|}{0.292}                                                     & 79.29                              \\ 
\textbf{FMNIST}                                        & \multicolumn{1}{c|}{7.921}                                                      & 43.86                             & \multicolumn{1}{c|}{\cellcolor{lightgreen}{4.986}}                                                      & \cellcolor{lightgreen}{92.99}                             & \multicolumn{1}{c|}{0.314}                                                     & 81.12                             \\ 
\textbf{CovType}                                       & \multicolumn{1}{c|}{64.746}                                                      & 65.35                             & \multicolumn{1}{c|}{\cellcolor{lightgreen}{6.532}}                                                      & \cellcolor{lightgreen}{99.22}                             & \multicolumn{1}{c|}{0.036}                                                    & 2.528                              \\
\textbf{TinyImages}                                    & \multicolumn{1}{c|}{10.501}                                                      & 27.62                            & \multicolumn{1}{c|}{\cellcolor{lightgreen}{3.074}}                                                      & \cellcolor{lightgreen}{73.11}                             & \multicolumn{1}{c|}{0.217}                                                     & 50.76                             \\ 
\textbf{Twitter}                                       & \multicolumn{1}{c|}{65.131}                                                      & 17.62                             & \multicolumn{1}{c|}{\cellcolor{lightgreen}{10.192}}                                                      & \cellcolor{lightgreen}{97.23}                             & \multicolumn{1}{c|}{0.042}                                                    & 00.78                               \\
\textbf{YearPred}                                      & \multicolumn{1}{c|}{58.845}                                                      & 22.97                          & \multicolumn{1}{c|}{\cellcolor{lightgreen}{7.299}}                                                      & \cellcolor{lightgreen}{90.48}                             & \multicolumn{1}{c|}{0.241}                                                     & 8.731                              \\ 
\textbf{SIFT}                                          & \multicolumn{1}{c|}{115.584}                                                      & 73.38                             & \multicolumn{1}{c|}{\cellcolor{lightgreen}{13.410}}                                                      & \cellcolor{lightgreen}{89.00}                             & \multicolumn{1}{c|}{1.020}                                                      & 57.02                             \\ 
\textbf{GIST}                                          & \multicolumn{1}{c|}{118.554}                                                       & 25.42                              & \multicolumn{1}{c|}{\cellcolor{lightgreen}{140.993}}                                                      & \cellcolor{lightgreen}{78.59}                              & \multicolumn{1}{c|}{10.896}                                                      & 52.84                              \\ 
\textbf{OpenI-ResNet}                                        & \multicolumn{1}{c|}{1684.21}   &                                                   56.26   &     \multicolumn{1}{c|}{\cellcolor{lightgreen}{869.72}}    &                                                   \cellcolor{lightgreen}{85.80}                & \multicolumn{1}{c|}{56.07}                                                                 & 58.37         \\ 
\textbf{OpenI-ConvNeXt}                                        & \multicolumn{1}{c|}{1379.24}   &                                                   61.48   &     \multicolumn{1}{c|}{\cellcolor{lightgreen}{2475.75}}    &                                                   \cellcolor{lightgreen}{90.20}                & \multicolumn{1}{c|}{171.90}                                                                 & 75.01         \\ \hline \hline
\end{tabular}%
}
\caption{Performance comparison of the FAISS implementation of multiple NNS approaches on benchmark datasets. The \textcolor{lightgreen}{highlighted cells} represent the approach having maximum R@10 amongst all the approaches.}
\label{tab:nns-faiss-results}
\end{table}

\section{Pseudo codes of the Indexing and \DenseLinkSearch{}}


\section*{Supplementary material}

**Algorithm 1** Indexing Algorithm

1: **function** BUILDINDEX($\mathcal{D}, \mathcal{K}_{index}$)
2:     ▷ Build index from vector data $\mathcal{D}$ by finding $\mathcal{K}_{index}$ nearest neighbors for each vector
3:     $\mathcal{N} = length(\mathcal{D})$     ▷ Number of vectors in data
4:     $\mathcal{I} = \text{list}()$     ▷ Initialize the index - a list of lists of linkss
5:     **for each** vector $v$ in $\mathcal{D}$ **do**
6:         $\mathcal{R}_v = +\infty$     ▷ Neighborhood radius, also known as near distance
7:         $\mathcal{C}_v = +\infty$     ▷ Distance to closest indexed vector
8:         $\mathcal{H}_v = \text{heap}(\mathcal{K}_{index})$     ▷ Initialize max-heap $\mathcal{H}_v$ of size $\mathcal{K}_{index}$ to hold links for vector $v$
9:         $\mathcal{L}_v = \text{list}()$     ▷ Initializing list $\mathcal{L}_v$ to hold links for vector $v$
10:        $\mathcal{I}_v = \text{list}()$     ▷ Initialize list $\mathcal{I}_v$ to hold finished index links for vector $v$
11:     **end for**
12:     $\mathcal{H} = \text{heap}(\mathcal{N})$     ▷ Initialize max-heap $\mathcal{H}$ of size $\mathcal{N}$ to hold $\mathcal{C}_v$ for unindexed vectors
13:     $v = \mathcal{H}.\text{pop}()$     ▷ Choose first vector as the root node
14:     **for** $n = 1$ **to** $\mathcal{N}$ **do**     ▷ For all $\mathcal{N}$ vectors
15:         **for each** link $l$ in $\mathcal{H}_v$ **do**     ▷ Traverse each near link for vector $v$
16:             $\mathcal{I}_v.\text{append}(l)$     ▷ Store longer links (AKA descend links) to index list for vector $v$
17:         **end for**
18:         CREATELINKS($v$)     ▷ Creating nearest neighbor links for vector $v$
19:         $v = \mathcal{H}.\text{pop}()$     ▷ Pop vector furthest from existing nodes to become next node
20:     **end for**
21:     **for each** vector $v$ in $\mathcal{D}$ **do**
22:         **for each** link $l$ in $\mathcal{H}_v$ **do**     ▷ Traverse each near link for vector $v$
23:             $\mathcal{I}_v.\text{append}(l)$     ▷ Store nearest neighbor links (AKA spread links) to index list for vector $v$
24:         **end for**
25:         $\mathcal{I}_v.\text{sort}()$     ▷ Sort links by increasing distance for vector $v$
26:         $\mathcal{I}.\text{append}(\mathcal{I}_v)$     ▷ Add the vector list $\mathcal{I}_v$ to full index $\mathcal{I}$
27:     **end for**
28:     **return** $\mathcal{I}$     ▷ Return index for each vector in data $\mathcal{D}$
29: **end function**

**Algorithm 2** Creating Links

1: **function** CREATELINKS($\mathcal{A}$)
2:     ▷ Create nearest neighbor links for vector $\mathcal{A}$, after which it will be considered indexed
3:     $\mathcal{P} = \text{COLLECTNEIGHBORS}(\mathcal{A})$     ▷ Collecting $2^{nd}$ neighbors for vector $\mathcal{A}$
4:     **for each** vector $\mathcal{B}$ in $\mathcal{P}$ **do**
5:         $d_{\mathcal{A},\mathcal{B}} = dist(\mathcal{A}, \mathcal{B})$     ▷ Compute distance between $\mathcal{A}$ and $\mathcal{B}$
6:         **if** $d_{\mathcal{A},\mathcal{B}} < \mathcal{R}_\mathcal{A}$ **or** $d_{\mathcal{A},\mathcal{B}} < \mathcal{R}_\mathcal{B}$ **then**     ▷ Only create a link if either endpoint considers the other near
7:             **if** $d_{\mathcal{A},\mathcal{B}} < \mathcal{C}_\mathcal{B}$ **then**     ▷ Check if $\mathcal{A}$ is nearest to $\mathcal{B}$
8:                 $\mathcal{C}_\mathcal{B} = d_{\mathcal{A},\mathcal{B}}$     ▷ Update distance to closest indexed vector
9:                 $\mathcal{H}.\text{heapify}()$     ▷ Update heap $\mathcal{H}$ which has $\mathcal{C}_\mathcal{B}$ as a key for one of its vectors
10:            **end if**
11:            ADDLINK($\mathcal{H}_\mathcal{A}, \mathcal{L}_\mathcal{A}, \mathcal{B}, d_{\mathcal{A},\mathcal{B}}$)     ▷ Add a link to $\mathcal{B}$ to vector $\mathcal{A}$'s heap or list
12:            ADDLINK($\mathcal{H}_\mathcal{B}, \mathcal{L}_\mathcal{B}, \mathcal{A}, d_{\mathcal{A},\mathcal{B}}$)     ▷ Add a link to $\mathcal{A}$ to vector $\mathcal{B}$'s heap or list
13:         **end if**
14:     **end for**
15:     **return**
16: **end function**

**Algorithm 3** Add link

```
1: function ADDLINK($\mathcal{H}_\mathcal{X}, \mathcal{L}_\mathcal{X}, \mathcal{Y}, d_{\mathcal{X},\mathcal{Y}}$)
2:     ▷ Add a link $\mathcal{X}$ to $\mathcal{Y}$ to vector $\mathcal{X}$'s heap $\mathcal{H}_\mathcal{X}$ or list $\mathcal{L}_\mathcal{X}$ based on the distance $d_{\mathcal{X},\mathcal{Y}}$
3:     if $d_{\mathcal{X},\mathcal{Y}} \geq \mathcal{R}_\mathcal{X}$ then                                              ▷ Check if $\mathcal{Y}$ is near $\mathcal{X}$
4:         $\mathcal{L}_\mathcal{X}$.append($\mathcal{Y}, d_{\mathcal{X},\mathcal{Y}}$)                              ▷ If not, append new link to list $\mathcal{L}_\mathcal{X}$
5:     else                                                                            ▷ If so, push new link into heap $\mathcal{H}_\mathcal{X}$
6:         if $\mathcal{H}_\mathcal{X}$.isFull() then                                  ▷ Check if heap $\mathcal{H}_\mathcal{X}$ is full
7:             $\mathcal{V}, d_{\mathcal{X},\mathcal{V}} = \mathcal{H}_\mathcal{X}$.pop()                           ▷ If so, pop old top link to vector $\mathcal{V}$
8:             if $d_{\mathcal{X},\mathcal{V}} \leq \mathcal{R}_\mathcal{V}$ then                                  ▷ Check if $\mathcal{V}$ considers $\mathcal{X}$ near
9:                 $\mathcal{L}_\mathcal{X}$.append($\mathcal{V}, d_{\mathcal{X},\mathcal{V}}$)                    ▷ If so, append old link to list $\mathcal{L}_\mathcal{X}$.
10:            else
11:                delete $\mathcal{V}, d_{\mathcal{X},\mathcal{Y}}$                  ▷ If not, neither endpoint considers other near, old link deleted
12:            end if
13:        end if
14:        $\mathcal{H}_\mathcal{X}$.push($\mathcal{Y}, d_{\mathcal{X},\mathcal{Y}}$)                              ▷ Push new link into heap $\mathcal{H}_\mathcal{X}$
15:        $\mathcal{R}_\mathcal{X} = \mathcal{H}_\mathcal{X}$.peek()                 ▷ Update neighborhood radius $\mathcal{R}_\mathcal{X}$ to new distance at top of heap $\mathcal{H}_\mathcal{X}$
16:    end if
17:    return
18: end function
```

**Algorithm 4** Collect Neighbors
```
 1: function COLLECTNEIGHBORS($\mathcal{A}$)
 2:     ▷ Collect the $2^{nd}$ neighbors of the vector $\mathcal{A}$
 3:     N2$^{nd}$= list()                                                         ▷ Initialize the $2^{nd}$ neighbors set
 4:     N1$^{st}_H$= list()                                               ▷ Initialize the $1^{st}$ neighbors from heap set
 5:     N1$^{st}_L$= list()                                                ▷ Initialize the $1^{st}$ neighbors from list set
 6:     for each link $l$ in $\mathcal{H}_\mathcal{A}$ do                            ▷ Traverse each link in heap $\mathcal{H}_\mathcal{A}$ of $\mathcal{A}$
 7:         $\mathcal{B}, d_{\mathcal{A},\mathcal{B}} = l$                                                 ▷ Extract vector and distance
 8:         N1$^{st}_H$.append($\mathcal{B}$)                                            ▷ Add the vector $\mathcal{B}$ into N1$^{st}_H$
 9:     end for
10:     for each link $l$ in $\mathcal{L}_\mathcal{A}$ do                             ▷ Traverse each link in list $\mathcal{L}_\mathcal{A}$ of $\mathcal{A}$
11:         $\mathcal{B}, d_{\mathcal{A},\mathcal{B}} = l$                                                 ▷ Extract vector and distance
12:         if $d_{\mathcal{A},\mathcal{B}} > \mathcal{R}_\mathcal{B}$ then                                   ▷ Check if $\mathcal{B}$ considers $\mathcal{A}$ near
13:             $\mathcal{L}_\mathcal{A}$.remove($l$)                       ▷ Neither endpoint considers other near, drop link
14:         else
15:             N1$^{st}_L$.append($\mathcal{B}$)                                        ▷ Add the vector $\mathcal{B}$ into N1$^{st}_L$
16:         end if
17:     end for
18:     for each vector $u$ in N1$^{st}_H$ do                                     ▷ For each vector in N1$^{st}_H$
19:         for each link $l$ in $\mathcal{H}_u$ do                               ▷ Traverse each link in heap $\mathcal{H}_u$ of vector $u$
20:             $v, d_{u,v} = l$                                                 ▷ Extract vector and distance
21:             N2$^{nd}$.append($v$)                                            ▷ Add the vector $u$ into N2$^{nd}$
22:         end for
23:         for each link $l$ in $\mathcal{L}_u$ do                                ▷ Traverse each link in list $\mathcal{L}_u$ of vector $u$
24:             $v, d_{u,v} = l$                                                 ▷ Extract vector and distance
25:             if $d_{u,v} > \mathcal{R}_v$ then          ▷ Check if distance $d_{u,v}$ is greater than vector $v$ near distance $\mathcal{R}_v$
26:                 $\mathcal{L}_u$.remove($l$)                   ▷ Neither endpoint considers other near, drop link
27:             else
28:                 N2$^{nd}$.append($v$)                                        ▷ Add the vector $v$ into N2$^{nd}$
29:             end if
30:         end for
31:     end for
32:     for each vector $u$ in N1$^{st}_L$ do                                     ▷ For each vector in N1$^{st}_L$
33:         for each link $l$ in $\mathcal{H}_u$ do                               ▷ Traverse each link in heap $\mathcal{H}_u$ of vector $u$
34:             $v, d_{u,v} = l$                                                 ▷ Extract vector and distance
35:             N2$^{nd}$.append($v$)                                            ▷ Add the vector $v$ into N2$^{nd}$
36:         end for
37:     end for
38:     for each vector $u$ in N2$^{nd}$ do                                       ▷ For each vector in N2$^{nd}$
39:         if N1$^{st}_H$.contains($u$) or N1$^{st}_L$.contains($u$) or $u == \mathcal{A}$ then
40:             N2$^{nd}$.remove($u$)                         ▷ Don't include $u$ if it is a $1^{st}$ neighbor or self
41:         end if
42:     end for
43:     return N2$^{nd}$
44: end function
```

**Algorithm 5** Search Algorithm

1: **function** DENSELINKSEARCH($q, \mathcal{D}, \mathcal{K}_{search}, \mathcal{I}$)
2:     ▷ Find the $\mathcal{K}_{search}$ nearest neighbors to query $q$ using index links $\mathcal{I}$
3:     $\mathcal{N} = length(\mathcal{D})$     ▷ Number of vectors in data and index
4:     $L = [+\infty] * \mathcal{N}$     ▷ Initialize lookup table to avoid repeating distance calculations
5:     $F = [False] * \mathcal{N}$     ▷ Initialize flags to avoid repeating traversals
6:     $\mathcal{H} = \text{heap}(\mathcal{K}_{search})$     ▷ Create results heap $\mathcal{H}$ of size $\mathcal{K}_{search}$ to hold nearest vectors
7:     $v = \mathcal{D}[0]$     ▷ Start at root vector
8:     $d_{v,q} = dist(v, q)$     ▷ Calculate distance between vector $v$ and query vector $q$
9:     $L[v] = d_{v,q}$     ▷ Store distance $d_{v,q}$ in lookup table
10:    $\mathcal{H}.\text{push}(v, d_{v,q})$     ▷ Push the vector $v$ into results heap $\mathcal{H}$
11:    $\mathcal{V}_C = v$     ▷ Initialize closest vector
12:    $D_C = d_{v,q}$     ▷ Initialize closest distance
13:    DESCEND($q, \mathcal{H}, \mathcal{I}$)     ▷ Call the DESCEND procedure
14:    **return** $\mathcal{H}$     ▷ Return heap containing nearest neighbors of query vector
15: **end function**

**Algorithm 6** Descend Stage

1: **function** DESCEND($q, \mathcal{H}, \mathcal{I}$)
2:     ▷ Descend along links of the closest vector to query vector $q$ and store results in heap $\mathcal{H}$
3:     $\mathcal{L} = \mathcal{I}[\mathcal{V}_C]$     ▷ Get links for the closest vector
4:     $\mathcal{V}_N, D_N = \text{LINKTRAVERSE}(q, \mathcal{V}_C, \mathcal{H}, \mathcal{L})$     ▷ Traverse links, return nearest vector and distance found
5:     **if** $D_N < D_C$ **then**     ▷ Check if the closest vector has changed
6:        $\mathcal{V}_C = \mathcal{V}_N$     ▷ Update the closest vector and distance
7:        $D_C = D_N$
8:        DESCEND($q, \mathcal{H}, \mathcal{I}$)     ▷ Repeat the DESCEND procedure
9:     **else**
10:       SPREAD($q, \mathcal{H}, \mathcal{I}$)     ▷ Switch to the SPREAD procedure
11:    **end if**
12:    **return**
13: **end function**

**Algorithm 7** Spread Stage

1: **function** SPREAD($q, \mathcal{H}, \mathcal{I}$)
2:     ▷ Spread along links of the $\mathcal{K}_{search}$ nearest vectors to query vector $q$ and store results in heap $\mathcal{H}$
3:     $D_L = \mathcal{H}.\text{peek}()$     ▷ Get limit distance from top of results heap
4:     **for each** vector $v$ **in** $\mathcal{H}$ **do**     ▷ For each vector $v$ in heap $\mathcal{H}$
5:        **if** $F[v] == False$ **then**     ▷ Skip vector if it has already been traversed
6:           $\mathcal{L} = \mathcal{I}[v]$     ▷ Get index links for the near vector $v$
7:           $\mathcal{V}_N, D_N = \text{LINKTRAVERSE}(q, v, \mathcal{H}, \mathcal{L})$     ▷ Traverse links, return nearest vector and distance found
8:           **if** $D_N < D_L$ **then**     ▷ Check if a new near vector was found
9:              **break**     ▷ New near vector found - break out of loop
10:          **end if**
11:        **end if**
12:    **end for**
13:    **if** $D_N < D_C$ **then**     ▷ Check if the closest vector has changed
14:        $\mathcal{V}_C = \mathcal{V}_N$     ▷ Update the closest vector and distance
15:        $D_C = D_N$
16:        DESCEND($q, \mathcal{H}, \mathcal{I}$)     ▷ Switch to the the DESCEND procedure
17:    **else if** $D_N < D_L$ **then**     ▷ Check if a new near vector has been found
18:        SPREAD($q, \mathcal{H}, \mathcal{I}$)     ▷ Repeat the SPREAD procedure
19:    **end if**
20:    **return**
21: **end function**

**Algorithm 8** Link Traverse
---
1: **function** LINKTRAVERSE($q, x, \mathcal{H}, \mathcal{L}$)
2:     ▷ Follow links in list $\mathcal{L}$ for vector $x$, calculating distance to query vector $q$, and storing results in heap $\mathcal{H}$
3:     $\mathcal{V}_N$ = **null**     ▷ Initialize nearest vector $\mathcal{V}_N$
4:     $D_N = +\infty$     ▷ Initialize nearest distance $D_N$
5:     $D_T = \mathcal{H}.\text{peek}()$     ▷ Get distance from top of results heap
6:     **for each** link $l$ **in** $\mathcal{L}$ **do**     ▷ For each link $l$ in list $\mathcal{L}$
7:        $v = l$     ▷ Extract other endpoint vector $v$ from link
8:        **if** $L[v] == +\infty$ **then**     ▷ Skip vector if distance has already been calculated
9:           $d_{v,q} = dist(v, q)$     ▷ Calculate distance between vector $v$ and query vector $q$
10:          $L[v] = d_{v,q}$     ▷ Store distance $d_{v,q}$ in lookup table
11:          **if** $d_{v,q} < D_T$ **then**     ▷ If vector $v$ is near enough
12:             $\mathcal{H}.\text{push}(v, d_{v,q})$     ▷ Update results heap and top distance
13:             $D_T = d_{v,q}$
14:          **end if**
15:          **if** $d_{v,q} < D_N$ **then**     ▷ If vector $v$ is nearest found so far
16:             $\mathcal{V}_N = v$     ▷ Update nearest vector and distance
17:             $D_N = d_{v,q}$
18:          **end if**
19:        **end if**
20:     **end for**
21:     $F[x] = True$     ▷ Set traverse flag
22:     **return** $\mathcal{V}_N, D_N$     ▷ Return nearest vector and distance
23: **end function**



\end{document}